\documentclass[10pt,twocolumn,letterpaper]{article}

\usepackage{cvpr}
\usepackage{times}
\usepackage{epsfig}
\usepackage{graphicx}
\usepackage{amsmath}
\usepackage{amssymb}
\usepackage[noend]{algpseudocode}
\usepackage{algorithmicx,algorithm}
\usepackage{subcaption}
\usepackage{booktabs}
\usepackage{multirow}

\usepackage{graphbox}
\usepackage{array,booktabs}

\newcommand{\tabincell}[2]{\begin{tabular}{@{}#1@{}}#2\end{tabular}}

\usepackage[pagebackref=true,breaklinks=true,letterpaper=true,colorlinks,bookmarks=false]{hyperref}

\cvprfinalcopy % *** Uncomment this line for the final submission
\def\cvprPaperID{} % *** Enter the CVPR Paper ID here

\ifcvprfinal\fi
\begin{document}

%%%%%%%%% TITLE
\title{PointGroup: Dual-Set Point Grouping for 3D Instance Segmentation} 

\author{Li Jiang$^{1}$\thanks{Equal Contribution.} \quad Hengshuang Zhao$^{1*}$ \quad Shaoshuai Shi$^{1}$ \quad Shu Liu$^{2}$ \quad Chi-Wing Fu$^{1}$ \quad Jiaya Jia$^{1,2}$\\
	$^{1}$The Chinese University of Hong Kong \quad $^{2}$SmartMore\\
	{\tt\small \{lijiang, hszhao, cwfu, leojia\}@cse.cuhk.edu.hk} \quad {\tt\small ssshi@ee.cuhk.edu.hk} \quad {\tt\small sliu@smartmore.com} 
}

\maketitle
%\thispagestyle{empty}

%%%%%%%%% ABSTRACT
\begin{abstract}
Instance segmentation is an important task for scene understanding. Compared to the fully-developed 2D, 3D instance segmentation for point clouds have much room to improve.
In this paper, we present PointGroup, a new end-to-end bottom-up architecture, specifically focused on better grouping the points by exploring the void space between objects. We design a two-branch network to extract point features and predict semantic labels and offsets, for shifting each point towards its respective instance centroid. A clustering component is  followed to utilize both the original and offset-shifted point coordinate sets, taking advantage of their complementary strength. Further, we formulate the ScoreNet to evaluate the candidate instances, followed by the Non-Maximum Suppression (NMS) to remove duplicates.
We conduct extensive experiments on two challenging datasets, ScanNet v2 and S3DIS, on which our method achieves the highest performance, 63.6\% and 64.0\%, compared to 54.9\% and 54.4\% achieved by former best solutions in terms of mAP with IoU threshold 0.5.
\end{abstract}

%%%%%%%%% BODY TEXT
\section{Introduction}

Instance segmentation is a fundamental and challenging task that requires to predict not only the semantic labels but also the instance IDs for every object in the scene. It has drawn much interest recently, given the potential applications for both outdoor and indoor environment regarding autonomous driving, robot navigation, to name a few.

Convolutional neural networks has boosted the performance of 2D instance segmentation \cite{dai2016instance,he2017mask,liu2018path,chen2019hybrid}. 
However, given unordered and unstructured 3D point clouds, 2D methods cannot be directly extended to 3D points and make the latter remains very challenging~\cite{wang2018sgpn,hou20193d,yang2019learning}. In this paper, we address the challenging 3D point cloud instance segmentation task by exploring the void space between 3D objects, along with the semantic information, to better segment individual objects.

\begin{figure}[t]
	\begin{center}
		\begin{tabular}{@{\hspace{0.0mm}}c@{\hspace{1.0mm}}c@{\hspace{0.0mm}}}
			\hspace*{-2mm}
			\includegraphics[width=0.53\linewidth]{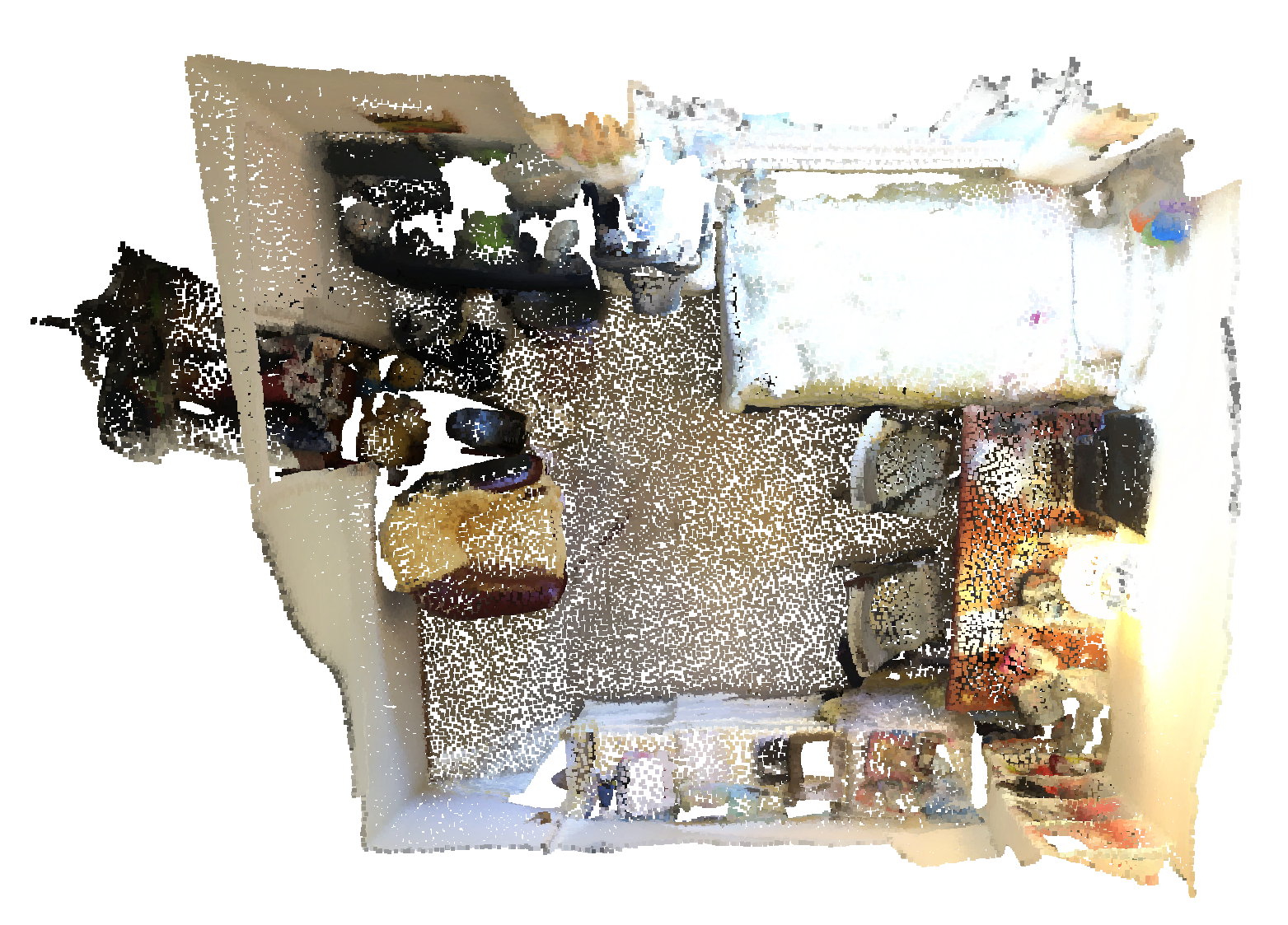}&
			\hspace*{-5mm}
			\includegraphics[width=0.53\linewidth]{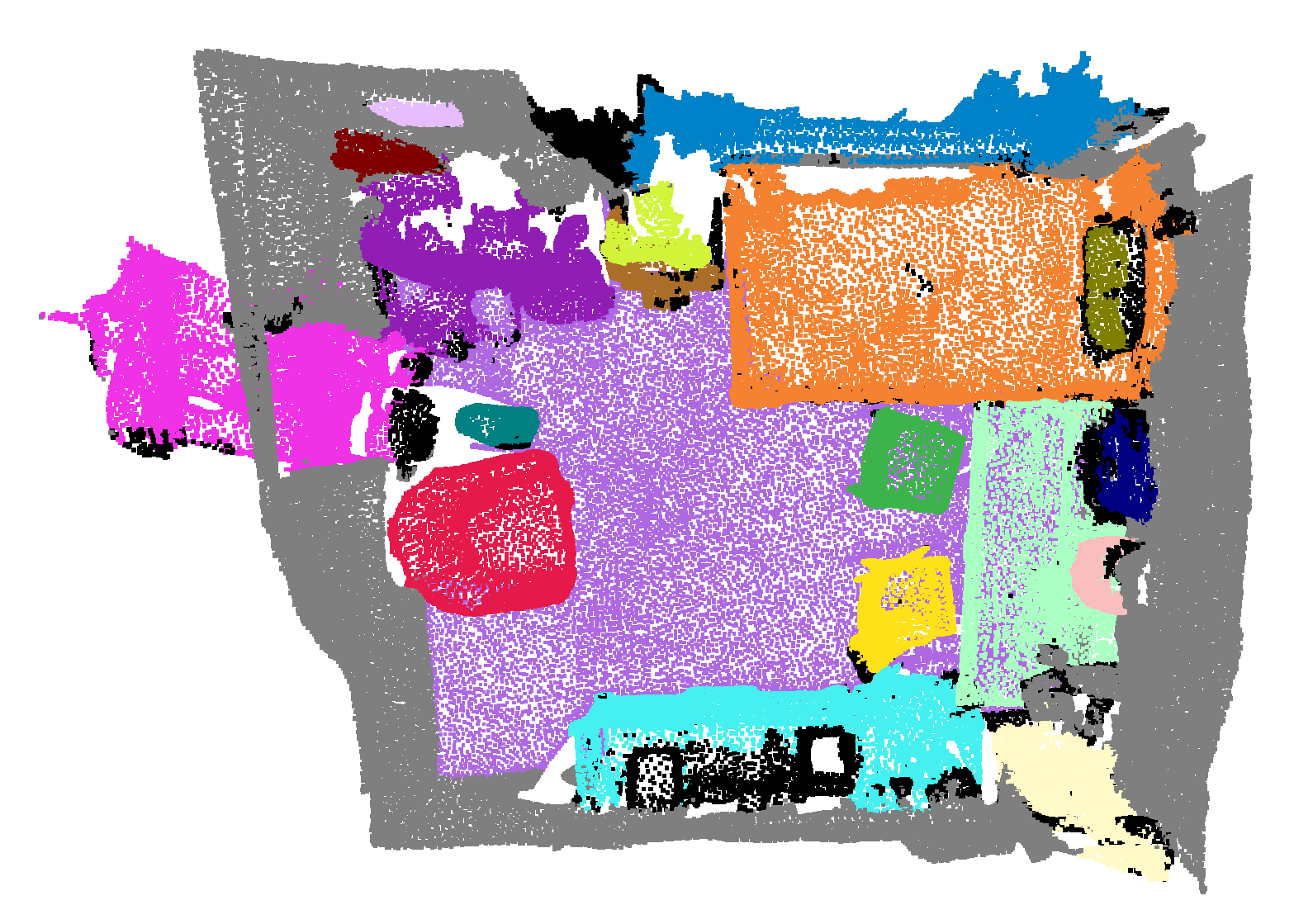} \\
			\hspace*{-2mm}
			Input &
			\hspace*{-5mm}
			Instance Prediction\\
		\end{tabular}
	\end{center}
	\vspace{-5mm}
	\caption{Example of 3D instance segmentation by our method from ScanNet v2.  
	Instances are in different colors.
}
	\vspace{-4mm}
	\label{fig:sample}
\end{figure}

Specifically, we design a bottom-up end-to-end framework named {\em PointGroup\/} for 3D instance segmentation, with the key target of better grouping of points. Our pipeline is to first extract per-point semantic prediction and conduct efficient point grouping to harvest candidate object instances. We utilize a semantic segmentation backbone to extract descriptive features and predict a semantic label for each point. 
Parallel to the segmentation head, we adopt an offset branch to learn a relative offset to bring each point to its respective ground-truth instance centroid. By this means, we shift points of the same object instance towards the same centroid and gather them closer, thus enabling better grouping of points into objects and separation of nearby objects of the same class.

With the predicted semantic labels and offsets, we then adopt a simple and yet effective algorithm to group points into clusters.
For each point, we take its coordinates as a reference, group it with nearby points of the same label, and expand the group progressively.
Importantly, we consider two coordinate sets in two separate passes -- the original point positions and those shifted by the predicted offsets. We call the process ``Dual-Set Point Grouping.'' The two types of results complement each other for accomplishing better performance.
Further, we design the ScoreNet to evaluate and pick candidate groups. Non-maximum suppression is finally adopted to remove duplicate predictions.

We conduct extensive experiments on the challenging ScanNet v2~\cite{dai2017scannet} and S3DIS~\cite{armeni2016s3dis} datasets.
PointGroup achieves the highest accuracy on both of them.
For ScanNet v2, our performance on the test set is 63.6\% in terms of mAP$_{50}$, which is 8.7\% higher than the former best solution~\cite{lahoud20193d}.
For S3DIS, we accomplish 64.0\% mAP$_{50}$, 69.6\% mPrec$_{50}$, and 69.2\% mRec$_{50}$, outperforming all previous approaches by a large margin. 

In summary, our contribution is threefold.
\begin{itemize}
	\vspace*{-0.8mm}
	\item We propose a bottom-up 3D instance segmentation framework, named {\em PointGroup\/}, to deal with the challenging 3D instance segmentation task.
	\vspace*{-0.8mm}
	\item We propose a point clustering method based on dual coordinate sets, \ie, the original and shifted sets. Along with the new ScoreNet, object instances can be better segmented out.
	\vspace*{-0.8mm}
	\item The proposed method achieves state-of-the-art results on various challenging datasets, demonstrating its effectiveness and generality.
\end{itemize}

%------------------------------------------------------------------------

%------------------------------------------------------------------------
\section{Related Work}
\paragraph{Deep Learning in 3D Scenes}
2D image pixels are in regular grids, thus can be naturally processed by convolutional neural networks~\cite{lecun1998gradient,krizhevsky2012imagenet,simonyan2014very,szegedy2015going,he2016deep}.
In contrast, 3D point clouds are unordered and scattered in 3D space, causing extra difficulty in point cloud scene understanding~\cite{qi2017pointnet2,shi2019pointrcnn}.

Several approaches handle data irregularity.
The Multi-Layer Perception (MLP)-style networks,~\eg, PointNet~\cite{qi2017pointnet,qi2017pointnet2}, directly apply MLP together with max-pooling to grab local and global structures in 3D. The learned features are then used for point cloud classification and segmentation.
Other approaches~\cite{dgcnn,wang2018pccn,zhao2019pointweb,wu2019pointconv,jiang2019hierarchical} enhance feature learning on local regions by dynamic context aggregation and attention modules. 

Besides working directly on the irregular input, several approaches transform the unordered point set to an ordered one to apply the convolution operations. PointCNN~\cite{pointcnn} learns the order transformation for points reweighting and permutation.
Some other approaches  \cite{maturana2015voxnet,song2016ssc,tchapmi2017segcloud,riegler2017octnet,graham20183d,choy20194d} align and voxelize point cloud to produce regular 3D ordered tensors for 3D convolution.
Multi-view strategies~\cite{qi2016volumetric, su15mvcnn, su20153dassisted} are also widely explored, where 3D point clouds are projected into 2D views for view-domain processing.

\vspace*{-3mm}
\paragraph{2D Instance Segmentation}
Instance segmentation aims to find the foreground objects in a scene and mark each object instance with a unique label.
Overall, there are two major lines.
The first is detection- or top-down-based, which directly detects object instances.
Early works~\cite{hariharan2014simultaneous,hariharan2015hypercolumns} use proposals from MCG~\cite{arbelaez2014multiscale} for feature extraction. Methods of \cite{dai2015convolutional,dai2016instance,hayder2017boundary} adopt pooled features for faster processing. 
Mask R-CNN~\cite{he2017mask} is widely known as an effective approach with the extra segmentation head in the detection framework, like Faster R-CNN~\cite{ren2015faster}.
Further works~\cite{liu2018path,chen2018masklab,chen2019hybrid} enhance the feature learning for instance segmentation.

The other line is segmentation- or bottom-up-based, where pixel-level semantic segmentation is performed followed by grouping of pixels to find object instances. 
Zhang~\etal~\cite{zhang2015monocular,zhang2016instance} utilize MRF for local patch merging.
Arnab and Torr~\cite{arnab2017pixelwise} use CRF.
Bai and Urtasun~\cite{bai2017deep} combine the classical watershed transform and deep learning to produce energy maps to distinguish among individual instances. Liu~\etal~\cite{liu2017sgn} employ a sequence of neural networks to construct objects from pixels.

\begin{figure*}
	\begin{center}
		\includegraphics[width=0.99\linewidth]{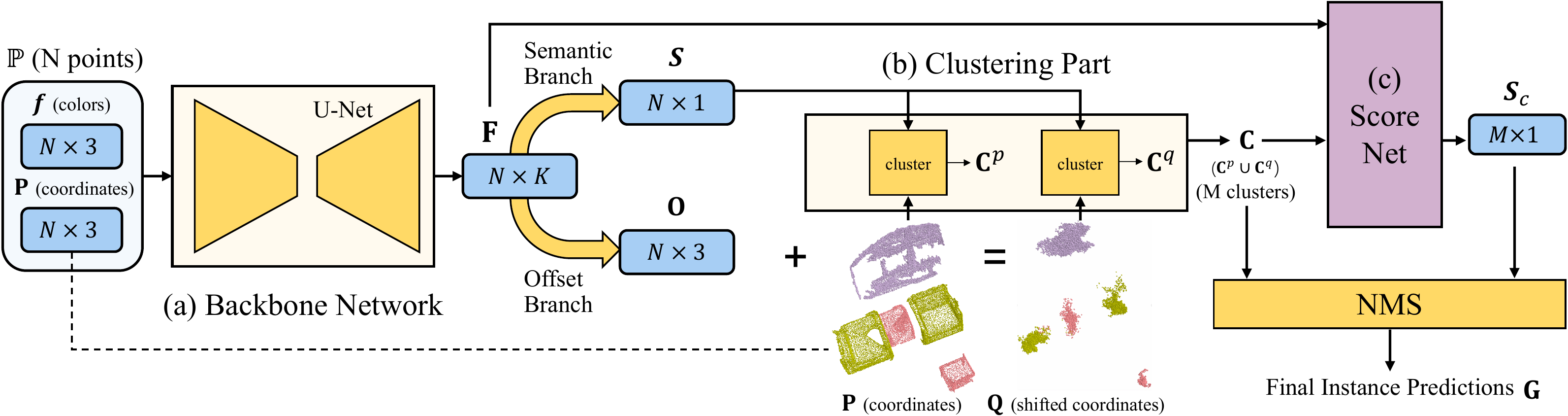}
	\end{center}
	\vspace{-4mm}
	\caption{Illustration of the network architecture. It has three main components -- (a) backbone network, (b) clustering part, and (c) ScoreNet.
		First, we use the backbone network to extract per-point features $\mathbf{F}$, followed by two branches to produce offset vectors $\mathbf{O} = \{o_i\}$ and semantic labels $\mathbf{S} = \{s_i\}$. 
		Then, we introduce a clustering method to group points into candidate clusters on dual coordinate sets, \ie, the original set $\mathbf{P}$ and the shifted $\mathbf{Q}$, which produce $\mathbf{C}^p$ and $\mathbf{C}^q$ respectively. 
		Lastly, we use ScoreNet to produce cluster scores $\mathbf{S}_c$. The set of color $\mathbf{f} = \{f_i\}$ serves as the input feature to the backbone.  \vspace{-1mm}}
	\label{fig:overview}
	
\end{figure*}

\vspace*{-3mm}
\paragraph{3D Instance Segmentation}
With available large-scale 3D labeled datasets~\cite{dai2017scannet,armeni2016s3dis}, instance segmentation of 3D point clouds becomes important. Similar to 2D cases, current 3D methods can also be grouped into two lines.

Detection-based methods extract 3D bounding boxes, and inside each box, utilize a mask learning branch to predict the object mask. Yang~\etal~\cite{yang2019learning} present the 3D-BoNet that directly predicts 3D bounding boxes and point-level masks simultaneously per instance.
Li~\etal~\cite{li2019GSPN} propose GSPN, which takes an analysis-by-synthesis strategy to generate proposals for instance segmentation.
Hou~\etal~\cite{hou20193d} combine multi-view RGB input with 3D geometry to jointly infer object bounding boxes and corresponding instance masks in an end-to-end manner.

Contrarily, segmentation-based methods predict the semantic labels, and utilize point embedding to group points into object instances.
Wang~\etal~\cite{wang2018sgpn} design SGPN by clustering points based on the semantic segmentation predicted by backbones such as PointNet++.
Liu and Furukawa~\cite{liu2019masc} predict both the semantic labels and affinity between adjacent voxels in different scales to group instances.
Phm~\etal~\cite{pham2019jsis3d} develop a multi-task learning framework with a multi-value CRF model to jointly reason over both the semantic and instance labels.
Wang~\etal~\cite{wang2019associatively} learn a semantic-aware point-level instance embedding to benefit learning of both the semantic and instance tasks.
Lahoud~\etal~\cite{lahoud20193d} introduce a
multi-task learning strategy where points of the same instance are grouped closer and different clusters are more separated from each other.

Different from the above methods, we present a new approach named {\em PointGroup} to tackle the 3D instance segmentation task. Our proposed model mainly contains two parts -- that is, (i) learning to group points into different clusters based on their semantic predictions in both the original coordinate space and shifted coordinate space, and (ii) ScoreNet to learn to predict the score for selecting proper clusters. The overall framework is differentiable. It can be jointly optimized and trained in an end-to-end manner.

\section{Our Method}

%%%%%%%%%%%%%%%%%%%%%%%%%%%%%%%%%%%%%%%%%%%%%%%%%%%%%%%%%%%%%%%%%%%%%%%

\subsection{Architecture Overview}
\label{sec:overview}
To obtain instance-level segmentation labels for 3D objects, we consider two problems.
The first is to separate the contents in the 3D space into individual objects, and the second is to determine the semantic label of each object.
Unlike 2D images, there is no view-occlusion problem in the 3D scenes, and objects scattered in 3D are usually naturally separated by void space.
Hence, we propose leveraging these characteristics of 3D objects to group 3D content into object instances according to the semantic information.
Fig.~\ref{fig:overview} overviews the architecture of our approach, which has three main components, \ie, the backbone, the point clustering part, and ScoreNet.

The input to the backbone network (Fig.~\ref{fig:overview}(a)) is a point set $\mathbb{P}$ of $N$ points. 
Each point has a color $f_i = (r_i, g_i, b_i)$ 
and 3D coordinates $p_i = (x_i, y_i, z_i)$, where $i \in \{1,...,N\}$.
The backbone extracts feature $F_i$ for each point.
We denote the output feature of the backbone as $\mathbf{F} = \{F_i\} \in \mathbb{R}^{N \times K}$, where $K$ is the number of channels.
We then feed $\mathbf{F}$ into two branches, one for semantic segmentation and the other for predicting a per-point offset vector to shift each point towards the centroid of its respective object instance.
Let $s_i$ and $o_i = (\Delta x_i, \Delta y_i, \Delta z_i)$ denote the predicted semantic label and offset vector of point $i$, respectively.

After obtaining the semantic labels, we begin to group points into instance clusters based on the void space between objects. 
In the point clustering part (Fig.~\ref{fig:overview}(b)), we introduce a clustering method to group points that are close to each other into the same cluster, if they have the same semantic label.
However,  
clustering directly based on the point coordinate set $\mathbf{P} = \{p_i\}$ may fail to separate same-category objects that are close to each other in the 3D space and mis-group them, for example, two pictures that hang side-by-side on the wall.

Thus, we 
use the learned offset $o_i$ to shift point $i$ towards its respective instance centroid and obtain the shifted coordinates $q_i = p_i + o_i \in \mathbb{R}^3$. 
For points belonging to the same object instance, different from $p_i$, the shifted coordinates $q_i$ clutter around the same centroid. So by clustering based on shifted coordinate set $\mathbf{Q} = \{q_i\}$, we separate nearby objects better, even though they have the same semantic labels.

However, for points near object boundary, the predicted offsets may not be accurate.
So, our clustering algorithm employs ``dual'' point coordinate sets, \ie, the original coordinates $\mathbf{P}$ and the shifted coordinates $\mathbf{Q}$.
We denote the clustering results $\mathbf{C}$ as the union of $\mathbf{C}^p = \{C^p_1, ..., C^p_{M_p}\}$ and $\mathbf{C}^q = \{C^q_1, ..., C^q_{M_q}\}$, which are the clusters discovered based on $\mathbf{P}$ and $\mathbf{Q}$, respectively. Here,  
$M_p$ and $M_q$ denote the number of clusters in $\mathbf{C}^p$ and $\mathbf{C}^q$, respectively, and 
$M = M_p + M_q$ denotes the total.

Lastly, we construct the ScoreNet (Fig.~\ref{fig:overview}(c)) to process the proposed point clusters $\mathbf{C} = \mathbf{C}^p \cup \mathbf{C}^q$ and produce a score per cluster proposal.
NMS is then applied to these proposals with the scores to generate final instance prediction.
In the following, we denote the instance predictions as $\mathbf{G} = \{G_1, ..., G_{M_{pred}} \} \subseteq \mathbf{C}$ and the ground-truth instances as $\mathbf{I} = \{I_1, ..., I_{M_{gt}}\}$.
Here, $G_i$ and $I_i$ are subsets of  $\mathbb{P}$, while $M_{pred}$ and $M_{gt}$ denote the number of instances in $\mathbf{G}$ and $\mathbf{I}$, respectively.
Also, we use $N^I_{i}$ and $N^G_{i}$ to represent the number of points in $I_i$ and $G_i$, respectively.

%%%%%%%%%%%%%%%%%%%%%%%%%%%%%%%%%%%%%%%%%%%%%%%%%%%%%%%%%%%%%%%%%%%%%%%
%\vspace*{-2mm}
\subsection{Backbone Network}
We may use any point feature extraction network to serve as the backbone network (Fig.~\ref{fig:overview}(a)).
In our implementation, we voxelize the points and follow the procedure of \cite{graham20183d} to construct a U-Net~\cite{li2019gs3d,ronneberger2015unet} with Submanifold Sparse Convolution (SSC) and Sparse Convolution (SC).
We then recover points from voxels to obtain the point-wise features. 
The contextual and geometric information is well extracted by the U-Net, which provides discriminative point-wise features $\mathbf{F}$ for subsequent processing.
Afterwards, we construct two branches based on the point-wise features $\mathbf{F}$ to predict semantic label $s_i$ and offset vector $o_i$ for each point.

%\vspace*{-3mm}
\paragraph{Semantic Segmentation Branch}
We apply an MLP to $\mathbf{F}$ to produce semantic scores $\mathbf{SC} = \{sc_1, ..., sc_N\} \in \mathbb{R}^{N \times N_{class}}$ for the $N$ points over the $N_{class}$ classes, and regularize the results by a cross entropy loss $L_{sem}$.
The predicted semantic label $s_i$ for point $i$ is the class with the maximum score,~\ie, $s_i = \text{argmax}(sc_i)$.

\vspace*{-3mm}
\paragraph{Offset Prediction Branch}
The offset branch encodes $\mathbf{F}$ to produce $N$ offset vectors $\mathbf{O} = \{o_1, ..., o_N\} \in \mathbb{R}^{N \times 3}$ for the $N$ points.
For points belonging to the same instance, we constrain their learned offsets by an $L_1$ regression loss as
\begin{equation}
	L_{o\_reg} = \frac{1}{\sum_i m_i}\sum_i ||o_i - (\hat{c}_i - p_i)|| \cdot m_i,
\end{equation}
where $\mathbf{m} = \{m_1, ..., m_N\}$ is a binary mask. $m_i = 1$ if point $i$ is on an instance and $m_i = 0$ otherwise.
$\hat{c}_i$ is the centroid of the instance that point $i$ belongs to,~\ie,
\begin{equation}
	\hat{c}_i = \frac{1}{N_{g(i)}^I}\sum\nolimits_{j \in I_{g(i)}} p_j,
\end{equation}
where $g(i)$ maps point $i$ to the index of its corresponding ground-truth instance,~\ie, the instance that contains point $i$. $N_{g(i)}^I$ is the number of points in instance $I_{g(i)}$.

\begin{figure*}
	\begin{subfigure}{.74\textwidth}
		\centering
		\includegraphics[width=0.99\linewidth]{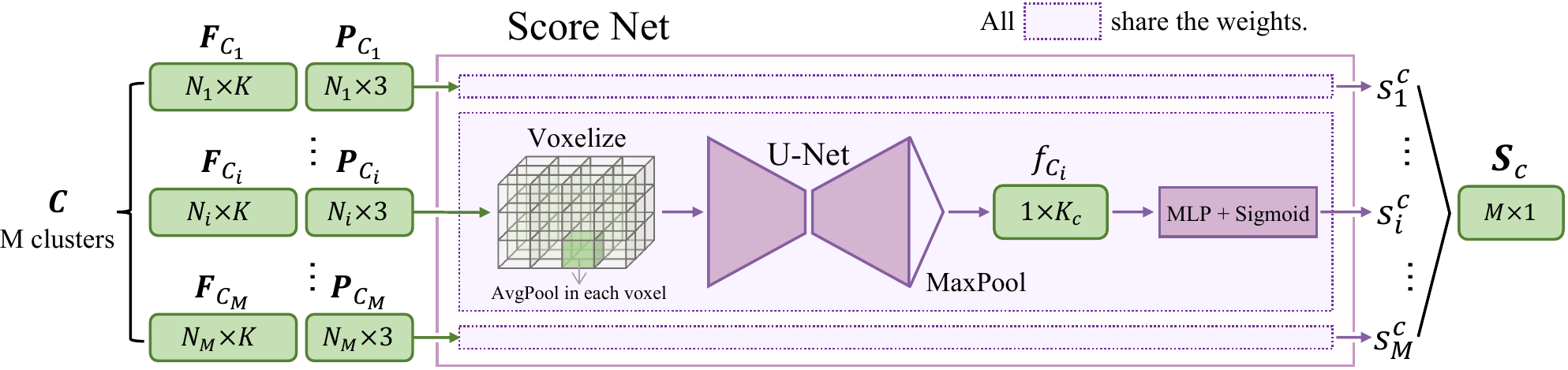}
		\caption{ScoreNet}
		\label{fig_scorenet}
	\end{subfigure}
	\begin{subfigure}{.24\textwidth}
		\begin{center}
			\includegraphics[width=0.9\linewidth]{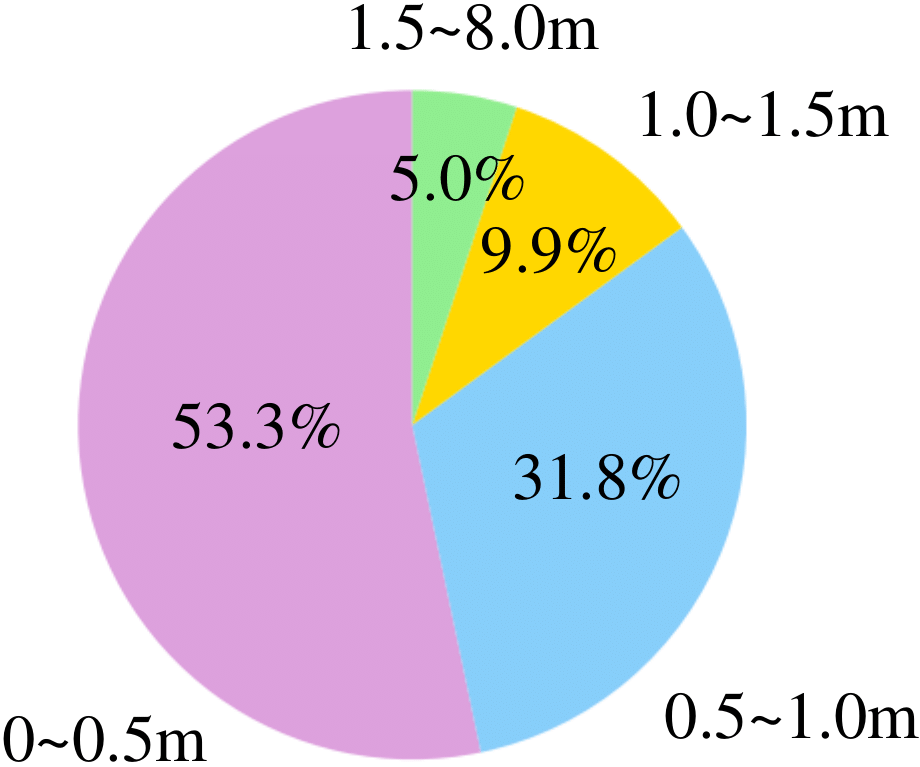}
		\end{center}
		\vspace*{-3mm}
		\caption{Distance Distribution}
		\label{fig_dist_dist}
	\end{subfigure}
	\vspace*{-1.5mm}
	\caption{(a) Structure of ScoreNet. (b) Distribution of distances from points to their respective instance centroids in the ScanNet dataset~\cite{dai2017scannet} (including the training and validation sets).}
	\vspace*{-1mm}
\end{figure*}

The above mechanism looks similar to the vote generation strategy in VoteNet~\cite{qi2019deep}.
However, rather than regressing the bounding boxes based on the votes of a few subsampled seed points, we predict an offset vector per point to gather the instance points around a common instance centroid, in order to better cluster relevant points into the same instance.
Also, we observe that the distances from points to their instance centroids usually have small values (0 to 1m). Fig.~\ref{fig_dist_dist} gives the statistical analysis on the distribution of such distances in the ScanNet dataset.
Considering diverse object sizes of different categories,
we find it is hard for the network to regress precise offsets, particularly for boundary points of large-size objects, since these points are relatively far from the instance centroids.
To address this issue, we formulate a direction loss to constrain the direction of predicted offset vectors. We follow~\cite{lahoud20193d} to define the loss as a means of minus cosine similarities,~\ie,
\begin{equation}
	L_{o\_dir} = - \frac{1}{\sum_i m_i} \sum_i \frac{o_i}{||o_i||_2} \cdot \frac{\hat{c}_i - p_i}{||\hat{c}_i - p_i||_2} \cdot m_i.
\end{equation}
Such loss is irrelevant to the offset vector norm and ensures that the points move towards their instance centroids.

%%%%%%%%%%%%%%%%%%%%%%%%%%%%%%%%%%%%%%%%%%%%%%%%%%%%%%%%%%%%%%%%%%%%%%%

\subsection{Clustering Algorithm}
Given the predicted semantic labels and offset vectors, we are ready to group the input points into instances. To this end, we introduce a simple and yet effective clustering algorithm. 
It is detailed in Algorithm~\ref{alg_cluster}. 

The core step of our algorithm is that for point $i$, we get points within the ball of radius $r$ centered at $x_i$ (the coordinate of point $i$) and group points with the same semantic labels as point $i$ into the same cluster.
Here, $r$ serves as a spatial constraint in the clustering, so that two intra-category objects at a distance larger than $r$ are not grouped.
Here, we use the breadth-first search to group points of the same instance into a cluster.
In our implementation, for points in the scene, neighboring points within an $r$-sphere can be found in parallel in advance of the clustering to boost speed.

As presented in Sec.~\ref{sec:overview}, we apply the clustering algorithm separately on the ``dual'' set,~\ie, the original coordinate set $\mathbf{P}$ and the shifted set $\mathbf{Q}$, to produce cluster sets $\mathbf{C}^p$ and $\mathbf{C}^q$.  
Clustering on $\mathbf{P}$ may mis-group nearby objects of the same class, while clustering on $\mathbf{Q}$ does not have this problem but may fail to handle the boundary points of large objects.
We collectively employ $\mathbf{P}$ and $\mathbf{Q}$ to find candidate clusters due to their complementary properties.
Analysis on the clustering performance of using $\mathbf{P}$ alone, $\mathbf{Q}$ alone, or both $\mathbf{P}$ and $\mathbf{Q}$ is presented in Sec.~\ref{sec_exp}.

\begin{algorithm}[t]
	\caption{Clustering algorithm.
		$N$ is the number of points.
		$M$ is the number of clusters found by the algorithm.}
	\label{alg_cluster}
	\hspace*{0.02in} {\bf Input:} clustering radius $r$; \\
	\hspace*{0.44in}
	cluster point number threshold $N_{\theta}$; \\
	\hspace*{0.44in}
	coordinates $\mathbf{X} = \{x_1, x_2, ..., x_N\}\in \mathbb{R}^{N \times 3}$; and \\
	\hspace*{0.44in}
	semantic labels $\mathbf{S} = \{s_1, ..., s_N\}\in \mathbb{R}^{N}$. \\
	\hspace*{0.02in} {\bf Output:} 
	clusters $\mathbf{C} = \{C_1, ..., C_M\}$.
	\begin{algorithmic}[1]
		\State initialize an array $v$ (visited) of length $N$ with all zeros
		\State initialize an empty cluster set $\mathbf{C}$
		\For{$i = 1$ to $N$}
		\If{$s_i$ is a stuff class (\eg, wall)}
		\State $v_i = 1$
		\EndIf
		\EndFor
		\For{$i = 1$ to $ N$}
		\If{$v_i == 0$}
		\State initialize an empty queue $Q$
		\State initialize an empty cluster $C$
		\State $v_i = 1$; $Q$.enqueue($i$); add $i$ to $C$
		\While{$Q$ is not empty}
		\State $k = Q$.dequeue()
		\For{$j \in [1, N]$ with $||x_j - x_k||_2 < r$}
		\If{$s_j == s_k$ and $v_j == 0$}
		\State $v_j = 1$; $Q$.enqueue($j$); add $j$ to $C$
		\EndIf
		\EndFor
		\EndWhile
		\If{number of points in $C > $$N_{\theta}$}
		\State add $C$ to $\mathbf{C}$
		\EndIf
		\EndIf
		\EndFor
		\State \Return $\mathbf{C}$
	\end{algorithmic}
\end{algorithm}

%%%%%%%%%%%%%%%%%%%%%%%%%%%%%%%%%%%%%%%%%%%%%%%%%%%%%%%%%%%%%%%%%%%%%%%

\subsection{ScoreNet}
\label{sec:score}
The input to ScoreNet is the set of candidate clusters $\mathbf{C} = \{C_1, ..., C_M\}$, where $M$ denotes the total number of candidate clusters, and $C_i$ denotes the $i$-th cluster. Also, we use $N_i$ to represent the number of points in $C_i$.
The goal of ScoreNet is to predict a score for each cluster to indicate the quality of the associated cluster proposal, so that we could precisely reserve the better clusters in NMS and thus combine strength of $\mathbf{C}^p$ and $\mathbf{C}^q$.

To start, for each cluster, we gather the point features from $\mathbf{F} \in \mathbb{R}^{N \times K}$ (the features extracted by the backbone) and form $\mathbf{F}_{C_i} = \{F_{h(C_i, 1)}, ..., F_{h(C_i, N_i)}\}$ for cluster $C_i$, where $h$ maps the point index in $C_i$ to corresponding point index in $\mathbb{P}$. Similarly, we express the coordinates for points in $C_i$ as $\mathbf{P}_{C_i} = \{p_{h(C_i, 1)}, ..., p_{h(C_i, N_i)}\}$. 

To better aggregate the cluster information, we take $\mathbf{F}_{C_i}$ and $\mathbf{P}_{C_i}$ as the initial features and coordinates, and voxelize the clusters the same way as we do at the beginning of the backbone network. The feature for each voxel is average-pooled from the initial features of points in that voxel. We then feed them into a small U-Net with SSC and SC to further encode the features. A cluster-aware max-pooling is then followed to produce a single cluster feature vector $f_{C_i} \in \mathbb{R}^{1 \times K_c}$ per cluster. The final cluster scores $\mathbf{S}_c = \{s^c_1, ..., s^c_M\} \in \mathbb{R}^{M}$ are obtained as
\begin{equation}
	\mathbf{S}_c = \text{Sigmoid}(\text{MLP}(\mathbf{F}_C)),
\end{equation}
where $\mathbf{F}_C = \{f_{C_1}, ..., f_{C_M}\} \in \mathbb{R}^{M \times K_c}$.
The structure of ScoreNet is illustrated in Fig.~\ref{fig_scorenet}.

Inspired by~\cite{li2019gs3d, jiang2018acquisition}, to reflect the quality of clusters in the scores, we use a soft label to replace a binary 0/1 label to supervise the predicted cluster score as
\begin{equation}
	\hat{s}^{c}_{i} = 
	\begin{cases}
		0     & {iou_i  <  \theta_l}\\
		1    & {iou_i > \theta_h}\\
		\frac{1}{\theta_h - \theta_l} \cdot (iou_i - \theta_l) & {otherwise}
	\end{cases},
\end{equation}
where $\theta_l$ and $\theta_h$ are empirically set to 0.25 and 0.75 respectively in our implementation, and 
$iou_i$ is the largest Intersection over Union (IoU) between cluster $C_i$ and ground-truth instances as
\begin{equation}
	iou_i = \text{max}\left(\left\{\text{IoU}(C_i, I_j) \mid  I_j \in \mathbf{I}\right\}\right).
\end{equation}
We then use the binary cross-entropy loss as our score loss, which is formulated as
\begin{equation}
	L_{c\_score} = - \frac{1}{M} \sum_{i=1}^{M} (\hat{s}^c_i log(s^c_i) + (1 - \hat{s}^c_i) log(1 - s^c_i)).
\end{equation}

%%%%%%%%%%%%%%%%%%%%%%%%%%%%%%%%%%%%%%%%%%%%%%%%%%%%%%%%%%%%%%%%%%%%%%%

\subsection{Network Training and Inference}

\paragraph{Training}
We train the whole framework in an end-to-end manner with the total loss as
\begin{equation}
	L = L_{sem} + L_{o\_dir} + L_{o\_reg} + L_{c\_score}.
\end{equation}

\paragraph{Inference}
In the inference process, we perform NMS on clusters $\mathbf{C}$ with predicted scores $\mathbf{S}_c$ to obtain the final instance predictions $\mathbf{G} \subseteq \mathbf{C}$. The IoU threshold is empirically set as 0.3. Since we cluster based on the semantic information, the semantic label of a cluster is exactly the category that the cluster points belong to.

\begin{table*}[th!]
	\centering
	\resizebox{\textwidth}{!}{
		\begin{tabular}{l|c|cccccccccccccccccc}
			%\hline
			\toprule[1pt]
			Method & Avg AP$_{50}$ &\rotatebox{90}{bathtub}&\rotatebox{90}{bed}&\rotatebox{90}{bookshe.}&\rotatebox{90}{cabinet}&\rotatebox{90}{chair}&\rotatebox{90}{counter}&\rotatebox{90}{curtain}&\rotatebox{90}{desk}&\rotatebox{90}{door}&\rotatebox{90}{otherfu.}&\rotatebox{90}{picture}&\rotatebox{90}{refrige.}&\rotatebox{90}{s. curtain}&\rotatebox{90}{sink}&\rotatebox{90}{sofa}&\rotatebox{90}{table}&\rotatebox{90}{toilet}&\rotatebox{90}{window}\\
			\hline
			SGPN~\cite{wang2018sgpn} & 0.143 & 0.208 & 0.390 & 0.169 & 0.065 & 0.275 & 0.029 & 0.069 & 0.000 & 0.087 & 0.043 & 0.014 & 0.027 & 0.000 & 0.112 & 0.351 & 0.168 & 0.438 & 0.138 \\
			3D-BEVIS~\cite{elich20193d} & 0.248 & 0.667 & 0.566 & 0.076 & 0.035 & 0.394 & 0.027 & 0.035 & 0.098 & 0.099 & 0.030 & 0.025 & 0.098 & 0.375 & 0.126 & 0.604 & 0.181 & 0.854 & 0.171 \\
			R-PointNet~\cite{li2019GSPN} &0.306 & 0.500 & 0.405 & 0.311 & 0.348 & 0.589 & 0.054 & 0.068 & 0.126 & 0.283 & 0.290 & 0.028 & 0.219 & 0.214 & 0.331 & 0.396 & 0.275 & 0.821 & 0.245 \\
			DPC~\cite{engelmann2019dilated} & 0.355 & 0.500 & 0.517 & 0.467 & 0.228 & 0.422 & \textbf{0.133} & 0.405 & 0.111 & 0.205 & 0.241 & 0.075 & 0.233 & 0.306 & 0.445 & 0.439 & 0.457 & 0.974 & 0.23 \\
			3D-SIS~\cite{hou20193d} & 0.382 & 1.000 & 0.432 & 0.245 & 0.190 & 0.577 & 0.013 & 0.263 & 0.033 & 0.320 & 0.240 & 0.075 & 0.422 & 0.857 & 0.117 & 0.699 & 0.271 & 0.883 & 0.235 \\
			MASC~\cite{liu2019masc} & 0.447 & 0.528 & 0.555 & 0.381 & 0.382 & 0.633 & 0.002 & 0.509 & 0.260 & 0.361 & 0.432 & 0.327 & 0.451 & 0.571 & 0.367 & 0.639 & 0.386 & 0.980 & 0.276 \\
			PanopticFusion~\cite{narita2019panopticfusion} & 0.478 & 0.667 & 0.712 & 0.595 & 0.259 & 0.550 & 0.000 & 0.613 & 0.175 & 0.250 & 0.434 & 0.437 & 0.411 & 0.857 & 0.485 & 0.591 & 0.267 & 0.944 & 0.35 \\
			3D-BoNet~\cite{yang2019learning} & 0.488 & 1.000 & 0.672 & 0.590 & 0.301 & 0.484 & 0.098 & 0.620 & 0.306 & 0.341 & 0.259 & 0.125 & 0.434 & 0.796 & 0.402 & 0.499 & 	0.513 & 0.909 & 0.439 \\
			MTML~\cite{lahoud20193d} & 0.549 & 1.000 & \textbf{0.807} & 0.588 & 0.327 & 0.647 & 0.004 & \textbf{0.815} & 0.180 & 0.418 & 0.364 & 0.182 & 0.445 & 1.000 & 0.442 & 0.688 & \textbf{0.571} & \textbf{1.000} & 0.396 \\
			\textbf{PointGroup (Ours)} & \textbf{0.636} & \textbf{1.000} & 0.765 & \textbf{0.624} & \textbf{0.505} & \textbf{0.797} & 0.116 & 0.696 & \textbf{0.384} & \textbf{0.441} & \textbf{0.559} & \textbf{0.476} & \textbf{0.596} & \textbf{1.000} & \textbf{0.666} & \textbf{0.756} & 0.556 & 0.997 & \textbf{0.513} \\
			\bottomrule[1pt]
		\end{tabular}
	}
	\vspace{-1mm}
	\caption{3D instance segmentation results on ScanNet v2 testing set with AP$_{50}$ scores. Our proposed PointGroup approach yields the highest average AP$_{50}$, outperforming all state-of-the-art methods by a large margin. All numbers are from the ScanNet benchmark on 15/11/2019.}
	\vspace{-2mm}
	\label{tab:scannet-test}
\end{table*}

\section{Experiments}
Our proposed PointGroup architecture is effective for instance segmentation of 3D point clouds. To demonstrate its effectiveness, we conduct extensive experiments on two challenging point cloud datasets, ScanNet v2~\cite{dai2017scannet} and S3DIS~\cite{armeni2016s3dis}. On both of them, we achieve state-of-the-art performance on the 3D instance segmentation task.

\subsection{Experimental Setting}

\vspace{2mm}
\paragraph{Datasets}
The ScanNet v2~\cite{dai2017scannet} dataset contains 1,613 scans with 3D object instance annotations. The dataset is split into training, validation, and testing sets, each with 1,201, 312, and 100 scans, respectively. 18 object categories are used for instance segmentation evaluation. For ablation studies, we train on the training set and report results on the validation set. To compare with other approaches, we train on the training set and report results on the testing set.

The S3DIS~\cite{armeni2016s3dis} dataset has 3D scans across six areas with 271 scenes in total. Each point is assigned one label out of 13 semantic classes. All the 13 classes are used in instance evaluation. Overall, we evaluate our model under two settings: (i) Area 5 is adopted for testing, whereas all the others are used for training; and (ii) six-fold cross validation that each area is treated as the testing set once.

\vspace*{-3mm}
\paragraph{Evaluation Metrics}
We use the widely-adopted evaluation metric -- mean average precision (mAP). Specifically, AP$_{25}$ and AP$_{50}$ denote the AP scores with IoU threshold set to 25\% and 50\%, respectively. Also, AP averages the scores with IoU threshold set from 50\% to 95\%, with a step size of 5\%. Besides, approaches of~\cite{wang2019associatively,yang2019learning} reported performance of mean precision (mPrec) and mean recall (mRec) on S3DIS, we also include these results for comparison.

\vspace*{-3mm}
\paragraph{Implementation Details}
We set the voxel size as 0.02m. 
In the clustering part, we set the clustering radius $r$ as 0.03m and the minimum cluster point number $N_\theta$ as 50. 
In the training process, we use the Adam solver with a base learning rate of 0.001. For each scene in the dataset, we set the maximum number of points as 250k, due to GPU memory limit. If the scene has more than 250k points, we randomly crop part of the scene and gradually adjust the crop size, according to the number of points in the cropped area. During the testing process, we feed the whole scene into the network without cropping.

Specifically, scenes in S3DIS have high point density. Some scenes are even with millions of points. Hence, for each S3DIS scene, we randomly sub-sample $\sim$1/4 points before each cropping.

\begin{table*}[th!]
	\centering
	\resizebox{\textwidth}{!}{
		\begin{tabular}{l|cc|ccccccccccccccccccc}
			\toprule[1pt]
			Method & Metric &  mean &\rotatebox{90}{bathtub}&\rotatebox{90}{bed}&\rotatebox{90}{bookshe.}&\rotatebox{90}{cabinet}&\rotatebox{90}{chair}&\rotatebox{90}{counter}&\rotatebox{90}{curtain}&\rotatebox{90}{desk}&\rotatebox{90}{door}&\rotatebox{90}{otherfu.}&\rotatebox{90}{picture}&\rotatebox{90}{refrige.}&\rotatebox{90}{s. curtain}&\rotatebox{90}{sink}&\rotatebox{90}{sofa}&\rotatebox{90}{table}&\rotatebox{90}{toilet}&\rotatebox{90}{window}\\
			\hline
			\multirow{3}{*}{Original $\mathbf{P}$} & AP & 0.283 & 0.414 & 0.327 & 0.244 & 0.167 & 0.493 & 0.083 & 0.269 & 0.089 & 0.193 & 0.286 & 0.205 & 0.207 & 0.373 & 0.226 & 0.361 & 0.251 & 0.684 & 0.231 \\
			& AP$_{50}$ & 0.507 & 0.692 & 0.647 & 0.481 & 0.347 & 0.685 & 0.231 & 0.508 & 0.308 & 0.384 & 0.453 & 0.359 & 0.301 & 0.632 & 0.537 & 0.660 & 0.531 & 0.961 & 0.413 \\
			& AP$_{25}$ & 0.659 & 0.840 & 0.764 & 0.597 & 0.496 & 0.791 & 0.588 & 0.614 & 0.686 & 0.529 & 0.600 & 0.432 & 0.401 & 0.660 & 0.775 & 0.777 & 0.721 & 0.995 & 0.601 \\
			\hline
			\multirow{3}{*}{Shifted $\mathbf{Q}$} & AP & 0.328 & 0.499 & 0.383 & 0.248 & 0.217 & 0.713 & 0.008 & 0.241 & 0.165 & 0.216 & 0.318 & 0.211 & 0.238 & 0.422 & 0.292 & 0.383 & 0.362 & 0.799 & 0.194 \\
			& AP$_{50}$ &0.529 & 0.738 & 0.694 & 0.550 & 0.435 & 0.884 & 0.035 & 0.389 & 0.410 & 0.413 & 0.501 & 0.363 & 0.366 & 0.617 & 0.590 & 0.648 & 0.571 & 0.948 & 0.375 \\
			& AP$_{25}$ & 0.677 & 0.863 & 0.795 & 0.699 & 0.617 & 0.931 & 0.426 & 0.541 & 0.697 & 0.538 & 0.623 & 0.446 & 0.366 & 0.765 & 0.826 & 0.848 & 0.669 & 0.999 & 0.533 \\
			\hline
			\multirow{3}{*}{Both $\mathbf{P}$ \& $\mathbf{Q}$} & AP & 0.348 & 0.597 & 0.376 & 0.267 & 0.253 & 0.712 & 0.069 & 0.266 & 0.140 & 0.229 & 0.339 & 0.208 & 0.246 & 0.416 & 0.298 & 0.434 & 0.385 & 0.758 & 0.275 \\
			& AP$_{50}$ & 0.569 & 0.805 & 0.696 & 0.549 & 0.481 & 0.877 & 0.224 & 0.449 & 0.416 & 0.420 & 0.530 & 0.377 & 0.372 & 0.644 & 0.611 & 0.715 & 0.629 & 0.983 & 0.462 \\
			& AP$_{25}$ & 0.713 & 0.865 & 0.795 & 0.744 & 0.673 & 0.925 & 0.648 & 0.616 & 0.741 & 0.548 & 0.654 & 0.482 & 0.383 & 0.711 & 0.828 & 0.851 & 0.742 & 1.000 & 0.636 \\
			\bottomrule[1pt]
		\end{tabular}
	}
	\caption{Ablation results using different coordinate sets on the ScanNet v2 validation set. Adopting both the original and shifted coordinates for clustering yields the best 3D instance segmentation performance.}
	\label{tab:coordinate}
\end{table*}

\subsection{Evaluation on ScanNet}

\subsubsection{Benchmark Results}
We first report performance of our PointGroup model on the testing set of ScanNet v2, as listed in Table~\ref{tab:scannet-test}. PointGroup accomplishes the highest AP$_{50}$ score of 63.6\%, outperforming all previous methods by a large margin. Compared with the former best solution~\cite{lahoud20193d}, which obtains 54.9\% AP$_{50}$ score, our result is 8.7\% higher (absolute) and 15.8\% better (relative). For detailed results on each category, PointGroup ranks the 1st place in 13 out of 18 classes in total.

\begin{figure*}
	\centering
	\begin{tabular}{@{\hspace{0.0mm}}c@{\hspace{1.0mm}}c@{\hspace{1.0mm}}c@{\hspace{1.0mm}}c@{\hspace{1.0mm}}c@{\hspace{0.0mm}}}
		\includegraphics[width=0.2\linewidth]{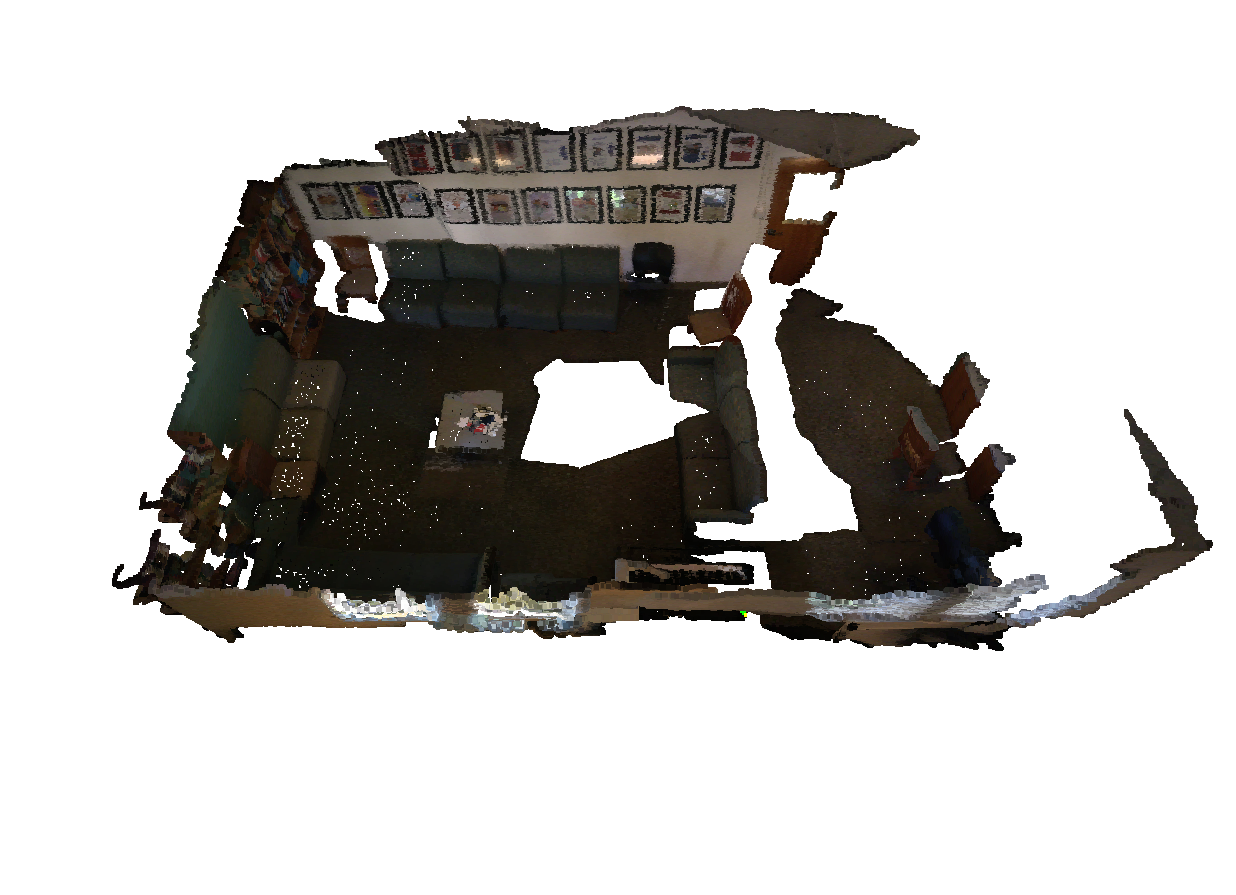}&
		\includegraphics[width=0.2\linewidth]{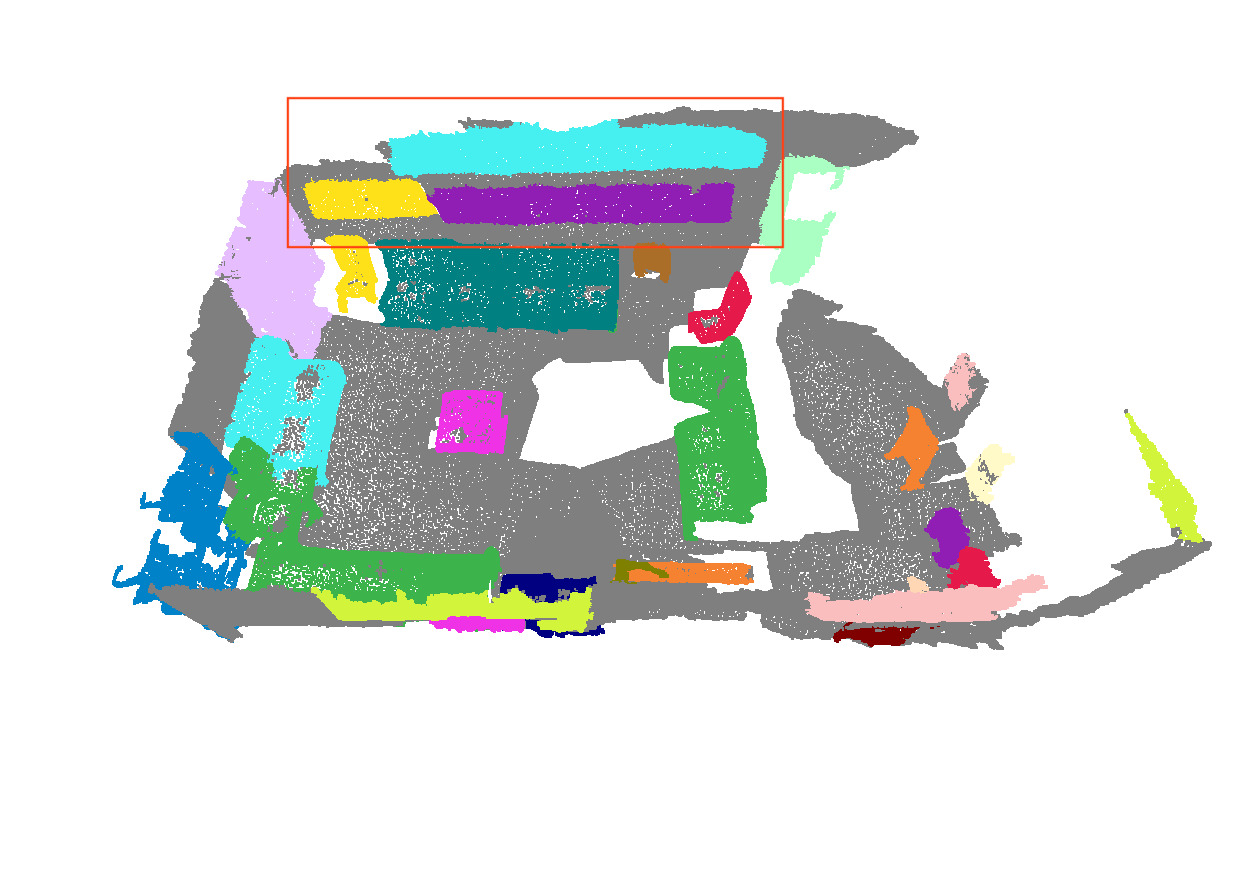}&
		\includegraphics[width=0.2\linewidth]{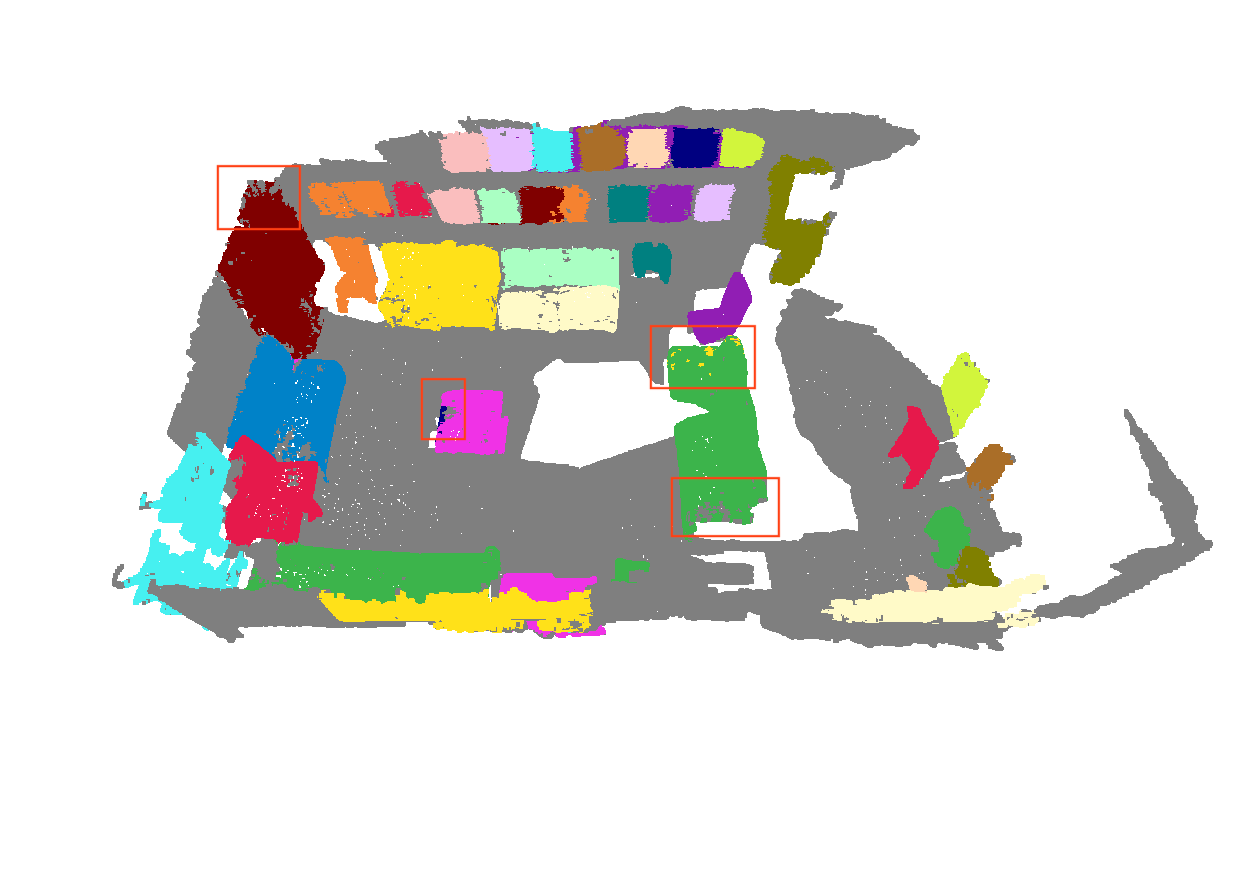}&
		\includegraphics[width=0.2\linewidth]{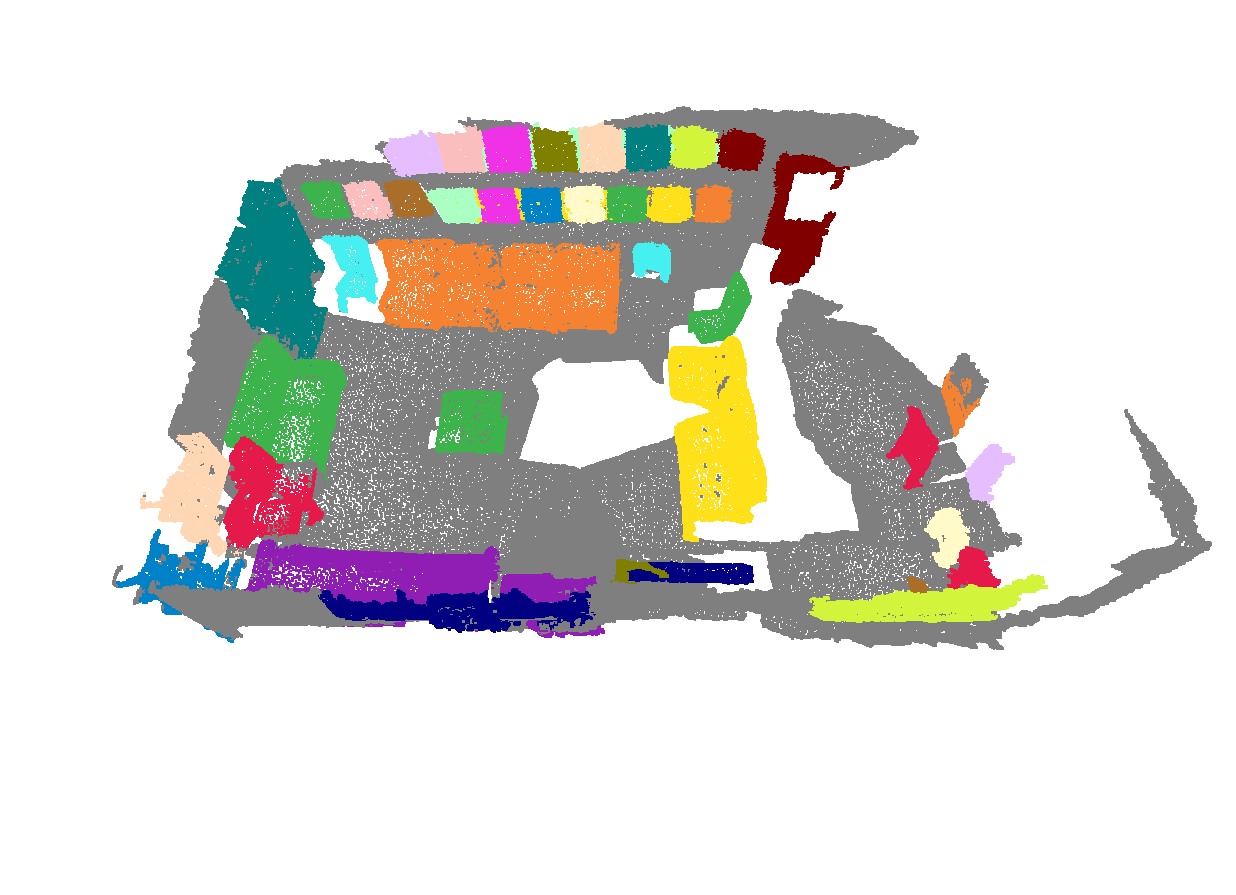}&
		\includegraphics[width=0.2\linewidth]{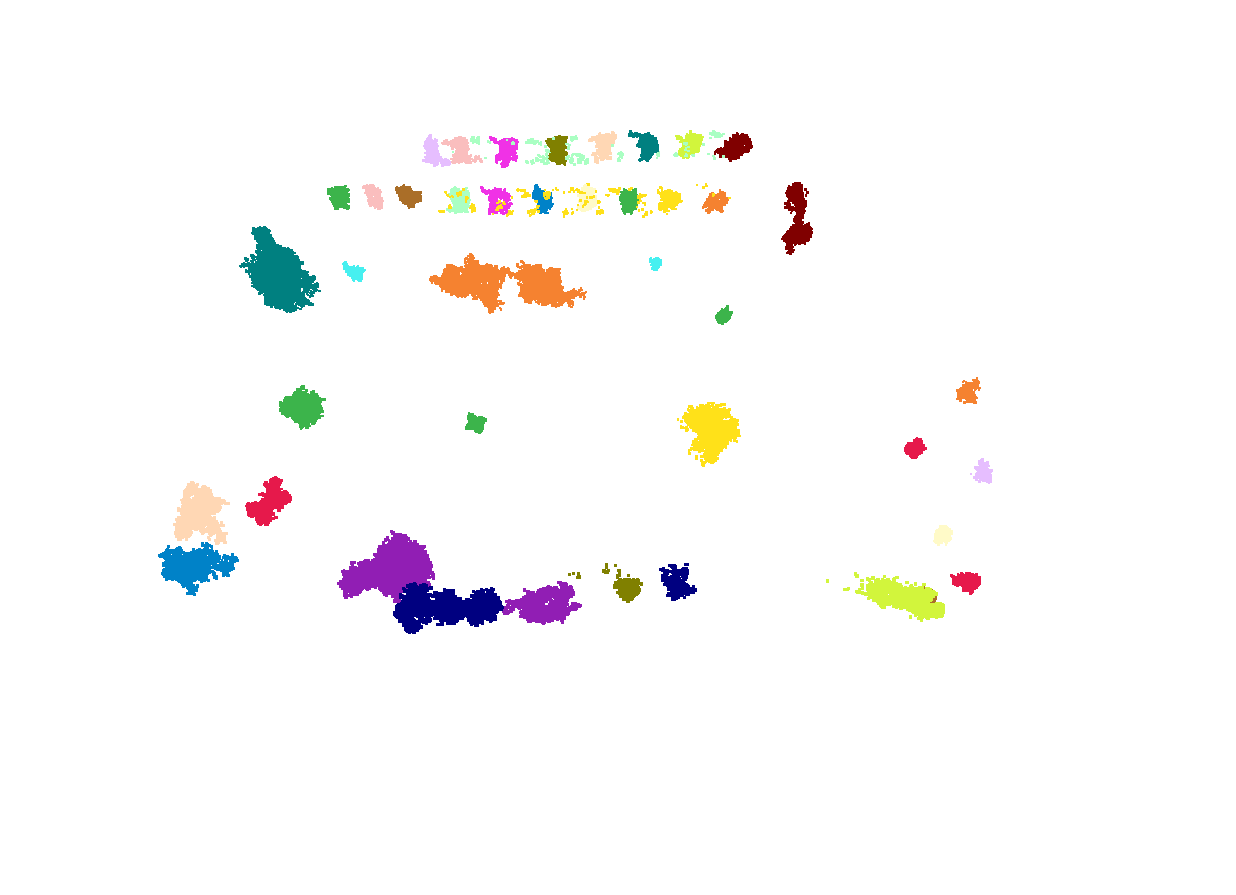}
		\vspace*{-5mm}
		\\
		Input &
		(i) $\mathbf{P}$ Only &
		(ii) $\mathbf{Q}$ Only &
		(iii) $\mathbf{P}$ and $\mathbf{Q}$ &
		Shifted Coord.
		\vspace*{-1mm}
		\\
	\end{tabular}
	\caption{Instance predictions produced by models trained with clustering on (i) $\mathbf{P}$ only,  (ii) shifted coordinates $\mathbf{Q}$ only, and (iii) both. The last column shows the predicted instances of (iii) represented with $\mathbf{Q}$, where stuff points are ignored.}
	\label{fig:ablation}
\end{figure*}

\subsubsection{Ablation Studies}
\label{sec_exp}
We conduct ablation studies on the ScanNet validation set to analyze the design and parameter choice in our PointGroup.

\vspace*{-3mm}
\paragraph{Clustering based on Different Coordinate Sets}
Table~\ref{tab:coordinate} shows the comparison using the original coordinates $\mathbf{P}$ alone, the shifted coordinates $\mathbf{Q}$ alone, and both $\mathbf{P}$ and $\mathbf{Q}$ in the clustering.
Clustering on points with $\mathbf{P}$ alone may mis-group two close objects with the same semantic label into the same instance. Hence, for categories, in which two objects are likely to be very close to each other (\eg, chairs and pictures), clustering on $\mathbf{P}$ alone does not perform well.  
Clustering on $\mathbf{Q}$ solves the problem in part by gathering instance points around the instance centroids and enlarging the space between clusters.  
However, due to inaccuracy in offset prediction, especially for boundary points of large objects (\eg, curtains and counters), clustering on $\mathbf{Q}$ alone does not perform perfectly.
  
Fig.~\ref{fig:ablation} shows the qualitative results with models trained with clusters from different coordinate sets -- (i) $\mathbf{P}$ only, (ii) $\mathbf{Q}$ only, and (iii) both $\mathbf{P}$ and $\mathbf{Q}$. We could observe that the problem in (i) is the mistakenly grouped pictures on the wall in one cluster. The case of (ii)
successfully separates the pictures into individual instances.  Nevertheless, it suffers from inaccuracy around the object boundary areas. The case of (iii) takes strength of both (i) and (ii). 
Clustering on dual point sets (both $\mathbf{P}$ and $\mathbf{Q}$) along with the precise scores from ScoreNet to select the final instance clusters, we combine the advantages of clustering on $\mathbf{P}$ and on $\mathbf{Q}$ to attain the best performance.

\begin{table}
	\vspace*{-0.5mm}
	\begin{center}
		\scalebox{0.85}[0.9]{
			\begin{tabular}{l | c c c}
				\hline
				Method & avg AP & avg AP$_{50}$ & avg AP$_{25}$ \\
				\hline
				$r$ = 2cm & 0.285 & 0.501 & 0.651 \\
				$r$ = 3cm & \textbf{0.348} & \textbf{0.569} & \textbf{0.713} \\
				$r$ = 4cm & 0.337 & 0.552 & 0.700 \\
				$r$ = 5cm & 0.342 & 0.552 & 0.699 \\
				\hline
			\end{tabular}
		}
	\end{center}
	\vspace{-4mm}
	\caption{Ablation results for clustering with different radii $r$ on the ScanNet v2 validation set.}
	\label{tab:radius}
\end{table}

\vspace*{-3mm}
\paragraph{Ablation on the Clustering Radius $r$}
We use different values of $r$ in the clustering algorithm. 
The performance varies as shown in Table~\ref{tab:radius}. A small $r$ is sensitive to point density. The scan for an object may have inconsistent point density in different parts. Clustering with such an $r$ may not be able to grow in low-density parts. On the contrary, a large $r$ increases the risk of grouping two nearby same-class objects into one. We empirically set $r$ to 0.03 (meter).

%\vspace*{-3mm}
\paragraph{Ablation for the ScoreNet}
We also ablate the ScoreNet, which is used to evaluate the quality of each candidate cluster (see Sec.\ref{sec:score}). 
Here, we directly use the output scores from ScoreNet to rank instances for calculating the AP. 

Apart from regressing the instance quality, an alternative way is to directly use the averaged semantic probability of the related instance category inside an instance as the quality confidence. 
By this means, the results in terms of AP/AP$_{50}$/AP$_{25}$ are 30.2/51.9/68.9(\%), which are worse than those with ScoreNet where results are 34.8/56.9/71.3(\%). This indicates that the proposed ScoreNet is vital and necessary for improving the instance segmentation results  by providing precise scores for NMS.

\subsubsection{Runtime Analysis}
Our method takes a whole scene as input per pass. Its runtime depends on the number of points and scene complexity. For runtime analysis, we sampled four scenes randomly from the ScanNet v2 validation set and tested them 100 times on a Titan Xp GPU to get an average runtime per scene. Table~\ref{tab:runtime} reports the runtime breakdown.
Clustering on $\mathbf{Q}$ (shifted) usually takes more time than clustering on $\mathbf{P}$ (original), as shifted points could have more neighbors.

\begin{table}
	\vspace*{-0.5mm}
	\setlength{\tabcolsep}{7pt}
	\footnotesize
	\begin{center}
		\scalebox{0.85}[0.9]{
			\begin{tabular}{ c | c | c||  c
					|@{\hspace{1mm}}c@{\hspace{1mm}}c@{\hspace{1mm}}|@{\hspace{1mm}}c@{\hspace{1mm}}c@{\hspace{1mm}}|
					c | c}
				\hline
				& \multirow{2}{*}{\tabincell{c}{\#Points}} &  \multirow{2}{*}{\tabincell{c}{Total\\Time}} &  \multirow{2}{*}{BB}  & \multicolumn{4}{c|}{Clustering on P and Q}  &  \multirow{2}{*}{SCN} &  \multirow{2}{*}{NMS} \\
				\cline{5-8}
				& &  &  & BQ$_p$ & CL$_p$ & BQ$_q$ & CL$_q$ & & \\
				\hline
				1 & 239,261 & 865 & 332 & 95 & 16 & 95 & 70 & 176 & 82 \\
				2 & 45,557 & 261 & 177 & 5 & 2 & 5 & 5 & 52 & 14  \\
				3 & 186,857 & 567 & 281 & 44 & 9 & 45 & 31 & 95 & 62 \\
				4 & 60,071 & 271 & 180 & 6 & 3 & 7 & 15 & 55 & 6\\
				\hline
				avg & 132,937 & {\bf 491} & 243 & 38 & 8 & 38 & 30 & 95 & 41 \\
				\hline
			\end{tabular}
		}
	\end{center}
	\vspace*{-4mm}
	\caption{Inference time (ms).
		BB denotes backbone $+$ two branches;
		BQ denotes ballquery;
		subscripts $p$ and $q$ denote clustering on $P$ and $Q$ respectively;
		CL denotes our clustering algorithm;
		and SCN denotes ScoreNet.}
	\label{tab:runtime}
\end{table}

\begin{table}
	\vspace{1mm}
	\begin{center}
		\scalebox{0.85}[0.9]{
			\begin{tabular}{ l | c c c}
				\hline
				Method & AP$_{50}$ & mPrec$_{50}$ & mRec$_{50}$ \\
				\hline
				SGPN$^\dag$~\cite{wang2018sgpn} & - & 0.360 & 0.287 \\
				ASIS$^\dag$~\cite{wang2019associatively} & - & 0.553 & 0.424 \\
				PointGroup$^\dag$ & \textbf{0.578} & \textbf{0.619} & \textbf{0.621} \\
				\hline
				SGPN$^\ddag$~\cite{wang2018sgpn} & 0.544 & 0.382 & 0.312 \\
				PartNet$^\ddag$~\cite{mo2019partnet} & - & 0.564 & 0.434 \\
				ASIS$^\ddag$~\cite{wang2019associatively} & - & 0.636 & 0.475 \\
				3D-BoNet$^\ddag$~\cite{yang2019learning} & - & 0.656 & 0.476 \\
				PointGroup$^\ddag$ & \textbf{0.640} & \textbf{0.696} & \textbf{0.692} \\
				\hline
			\end{tabular}
		}
	\end{center}
	\vspace{-4mm}
	\caption{Instance segmentation results on the S3DIS validation set.  Methods marked with $\dag$ are evaluated on Area 5; those marked with $\ddag$ are on the 6-fold cross validation. 
	}
	\label{tab:s3dis-compare}
\end{table}

\begin{figure*}[th!]
	\vspace{-2mm}
	\centering
	\begin{tabular}{@{\hspace{0.0mm}}c@{\hspace{1.0mm}}c@{\hspace{1.0mm}}c@{\hspace{1.0mm}}c@{\hspace{1.0mm}}c@{\hspace{0.0mm}}}
		\includegraphics[width=0.19\linewidth]{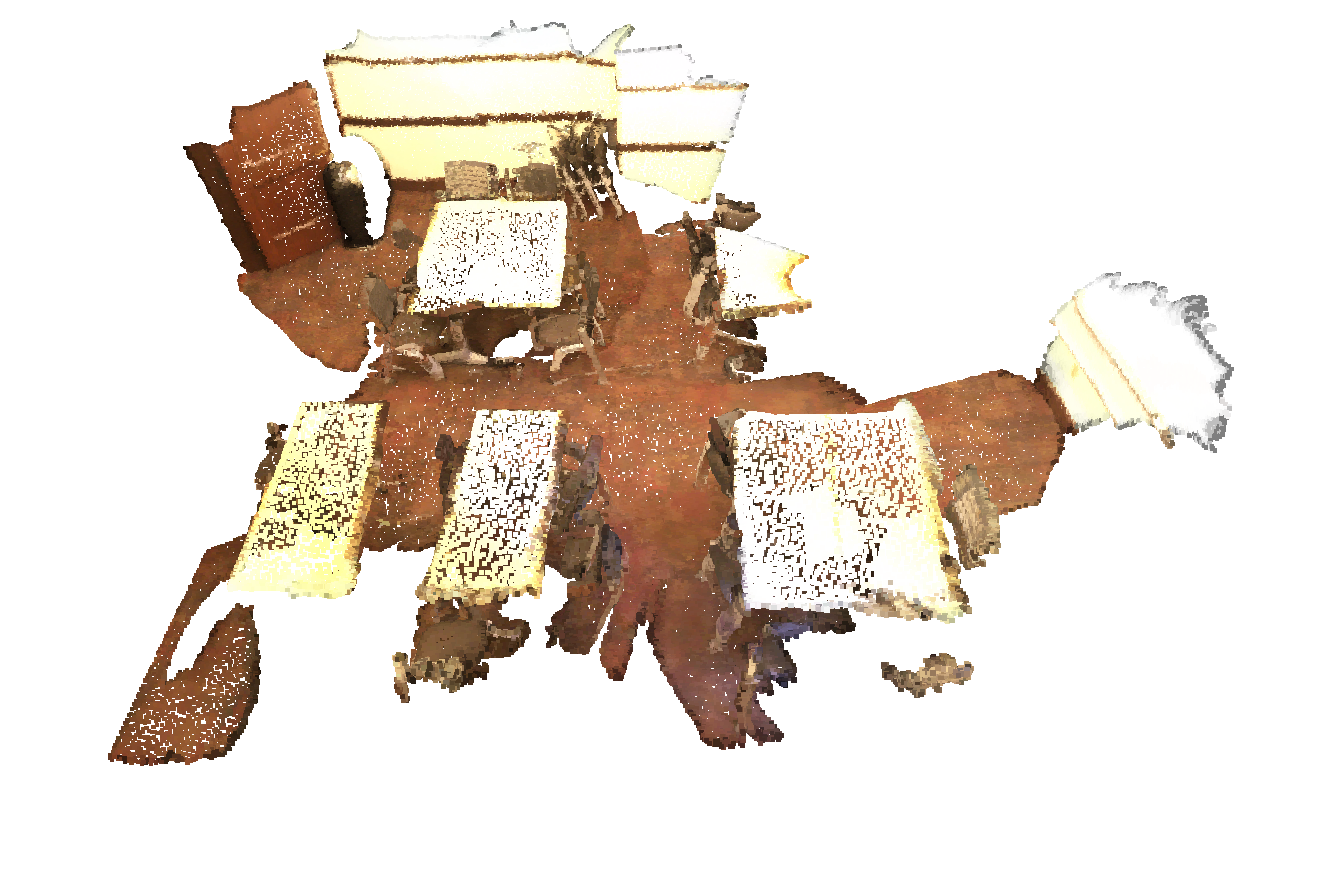}&
		\includegraphics[width=0.19\linewidth]{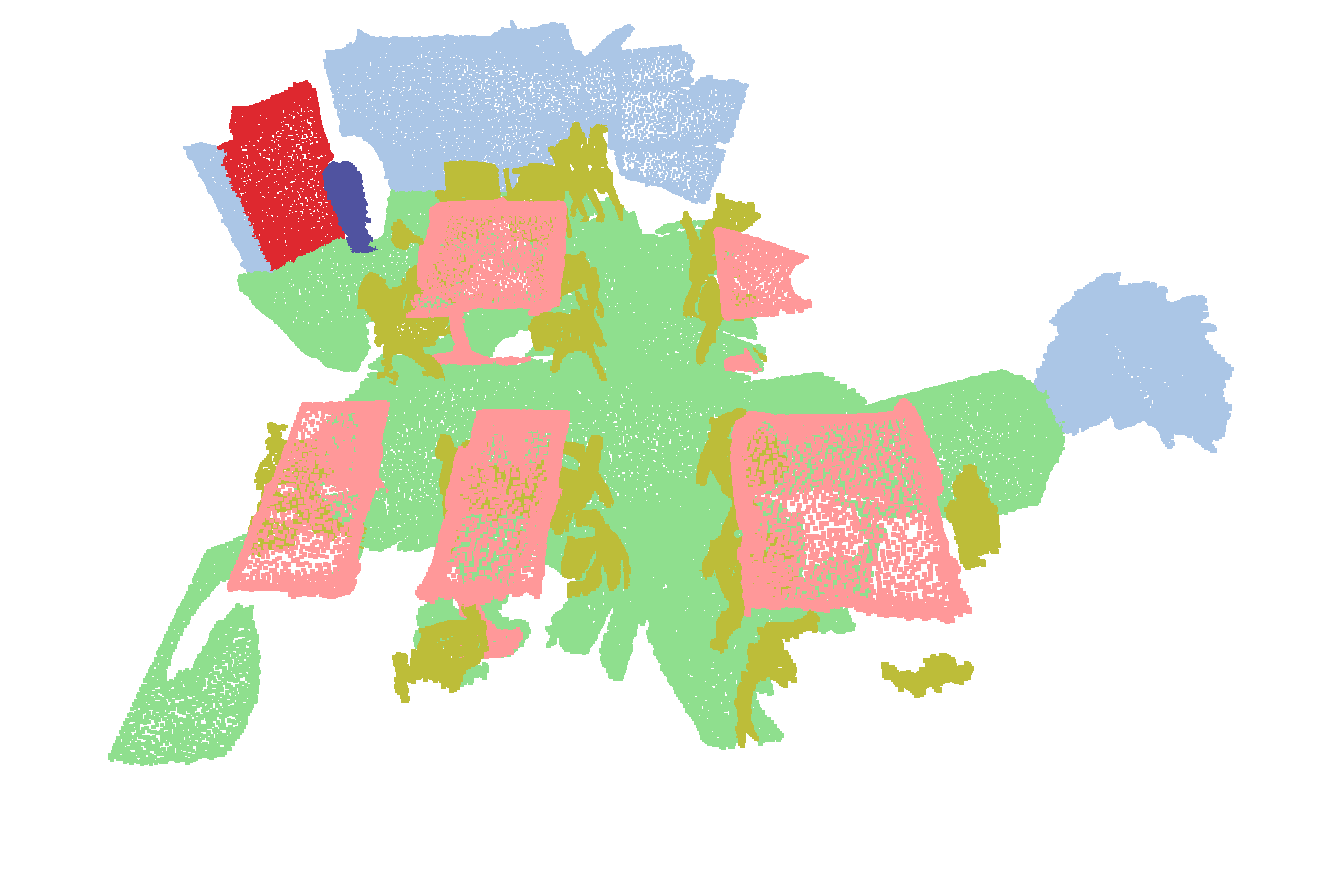}&
		\includegraphics[width=0.19\linewidth]{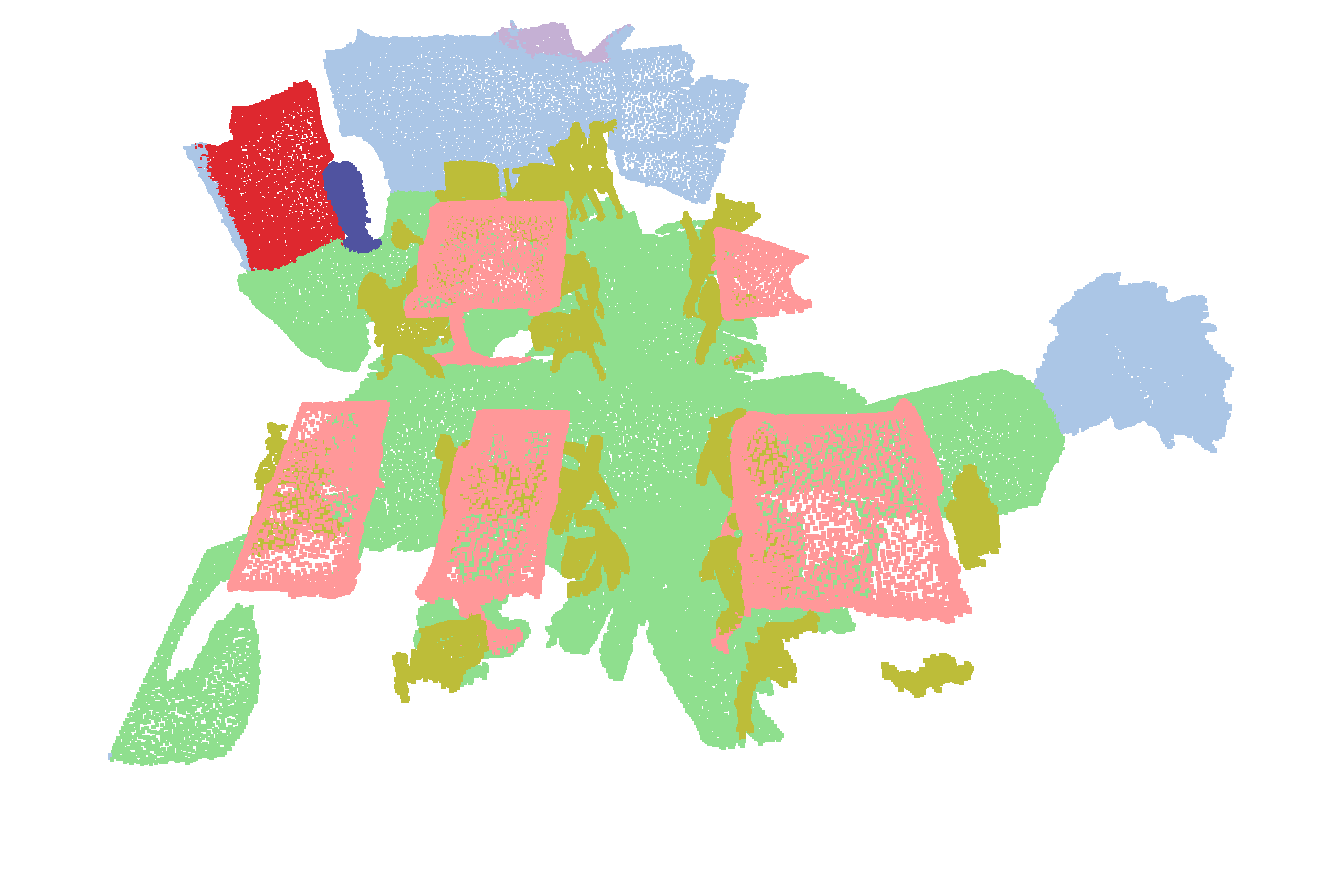}&
		\includegraphics[width=0.19\linewidth]{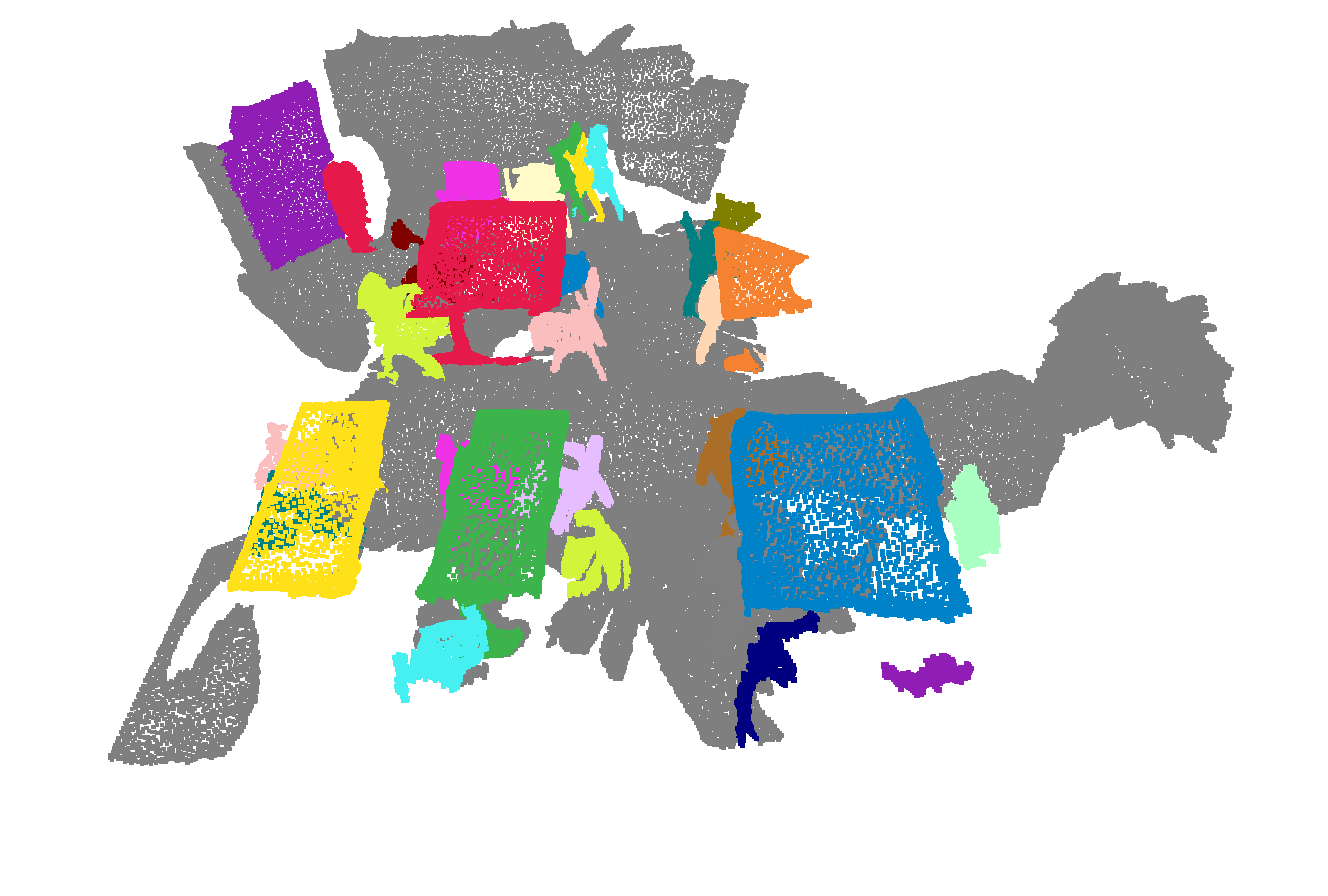}&
		\includegraphics[width=0.19\linewidth]{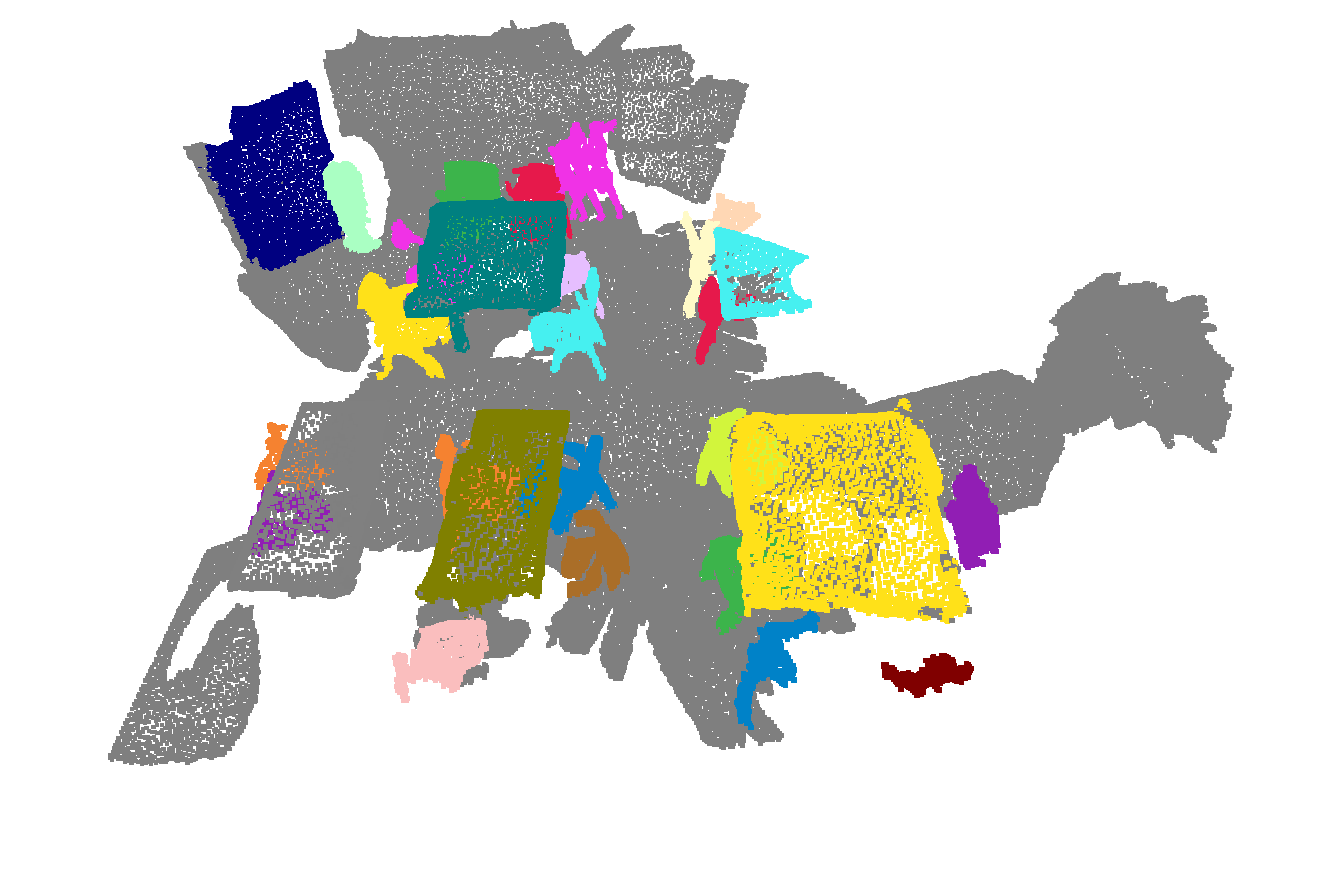}\\
		\includegraphics[width=0.19\linewidth, height=2.9cm]{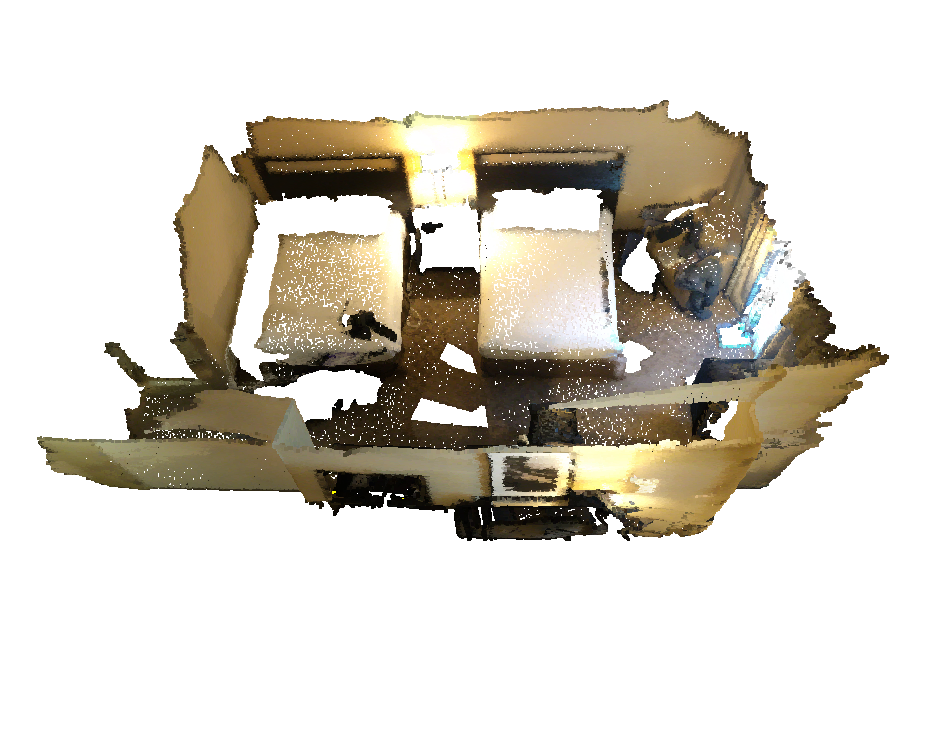}&
		\includegraphics[width=0.19\linewidth, height=2.9cm]{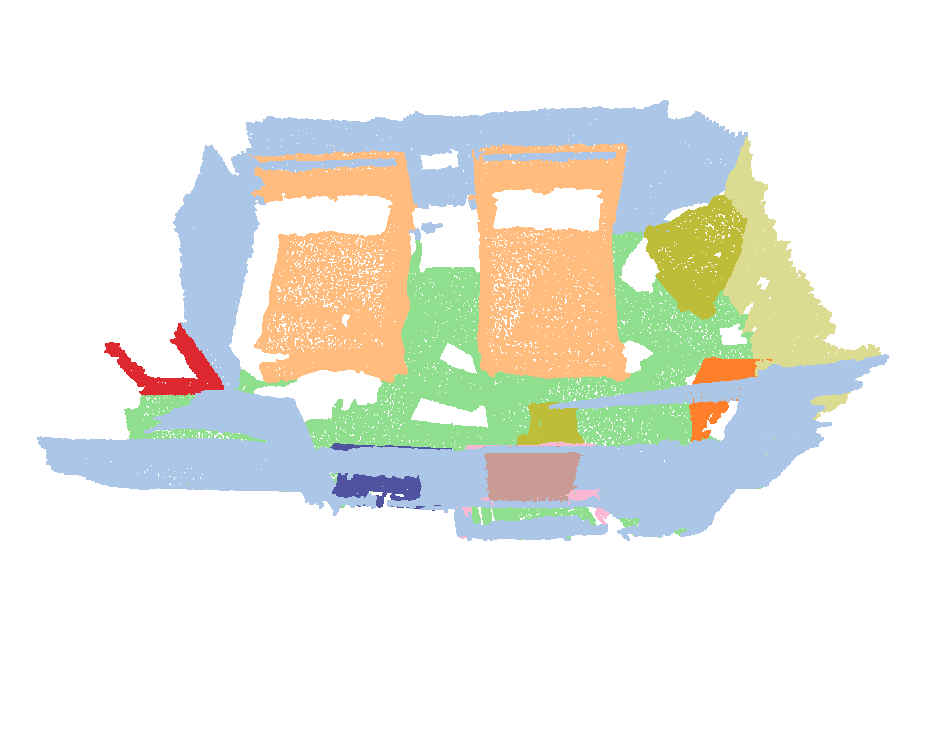}&
		\includegraphics[width=0.19\linewidth, height=2.9cm]{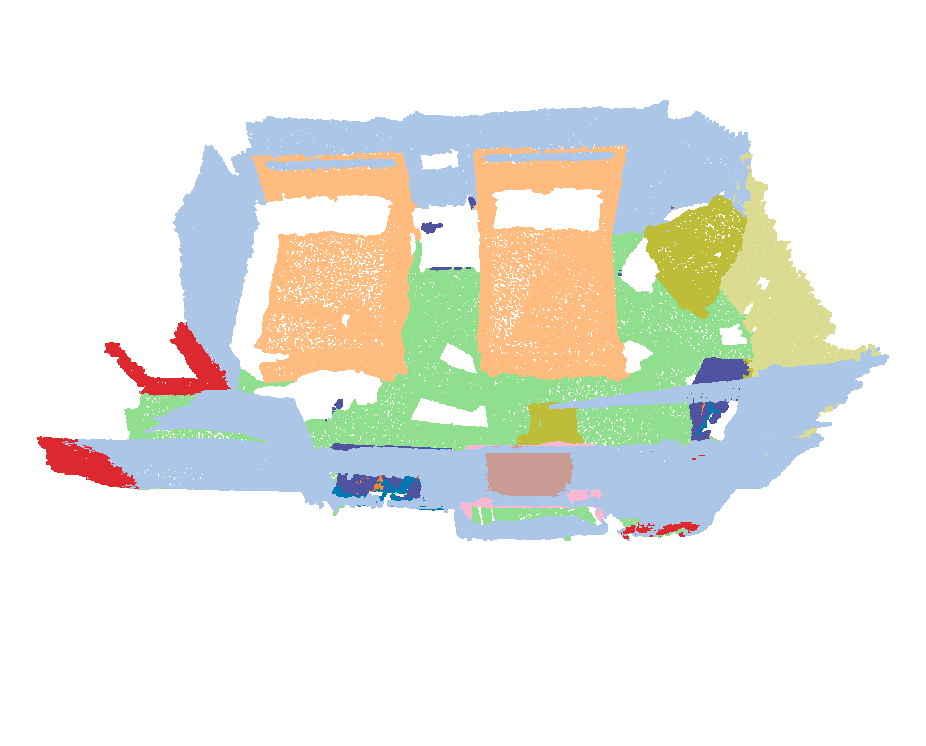}&
		\includegraphics[width=0.19\linewidth, height=2.9cm]{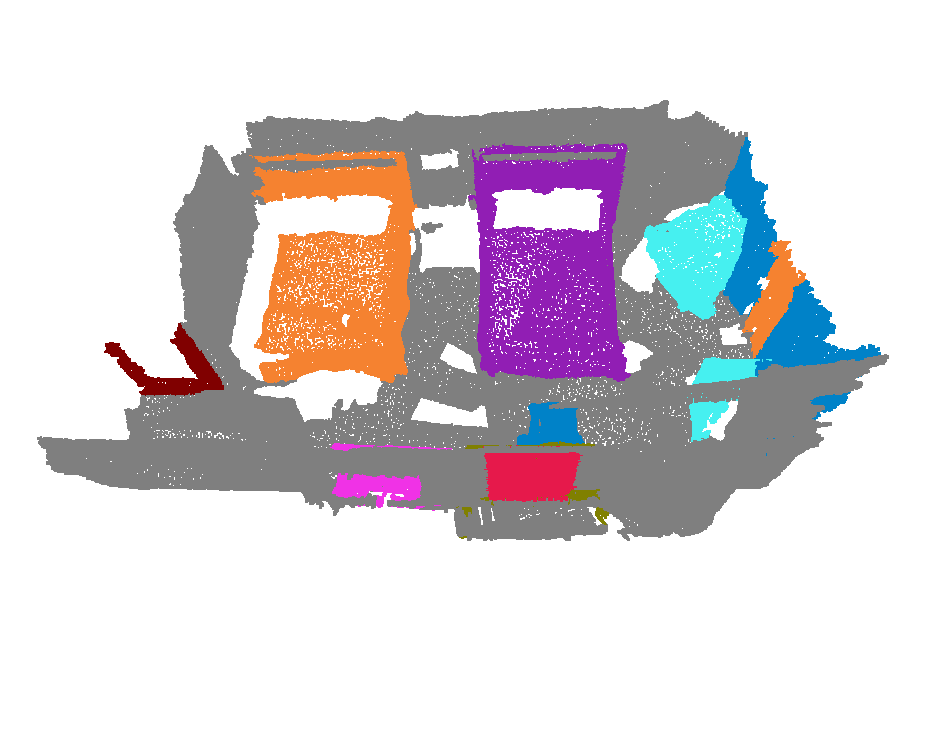}&
		\includegraphics[width=0.19\linewidth, height=2.9cm]{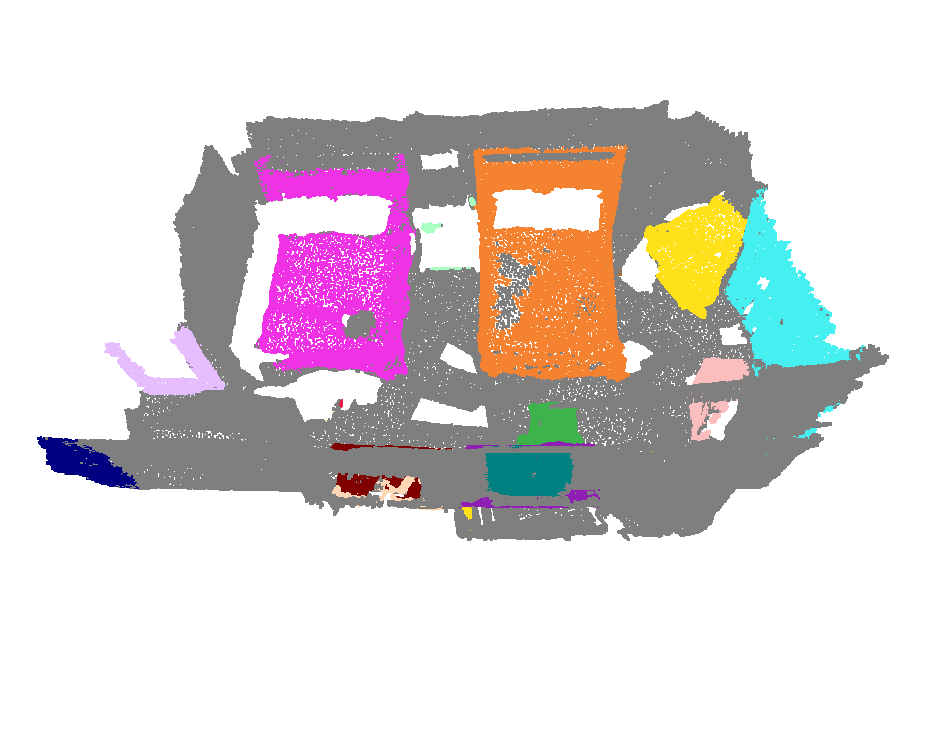}\\
		\includegraphics[width=0.19\linewidth, height=2.3cm]{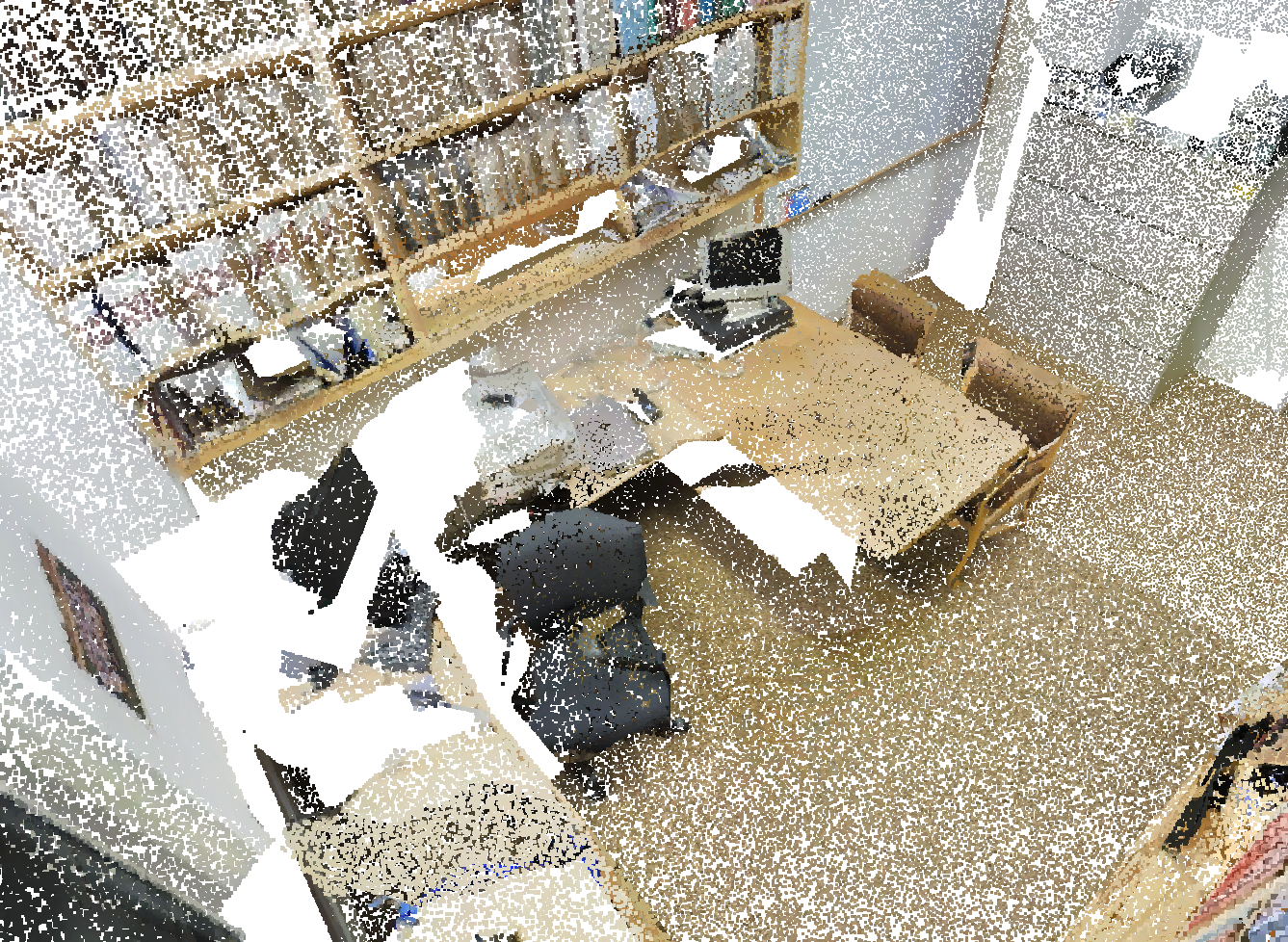}&
		\includegraphics[width=0.19\linewidth, height=2.3cm]{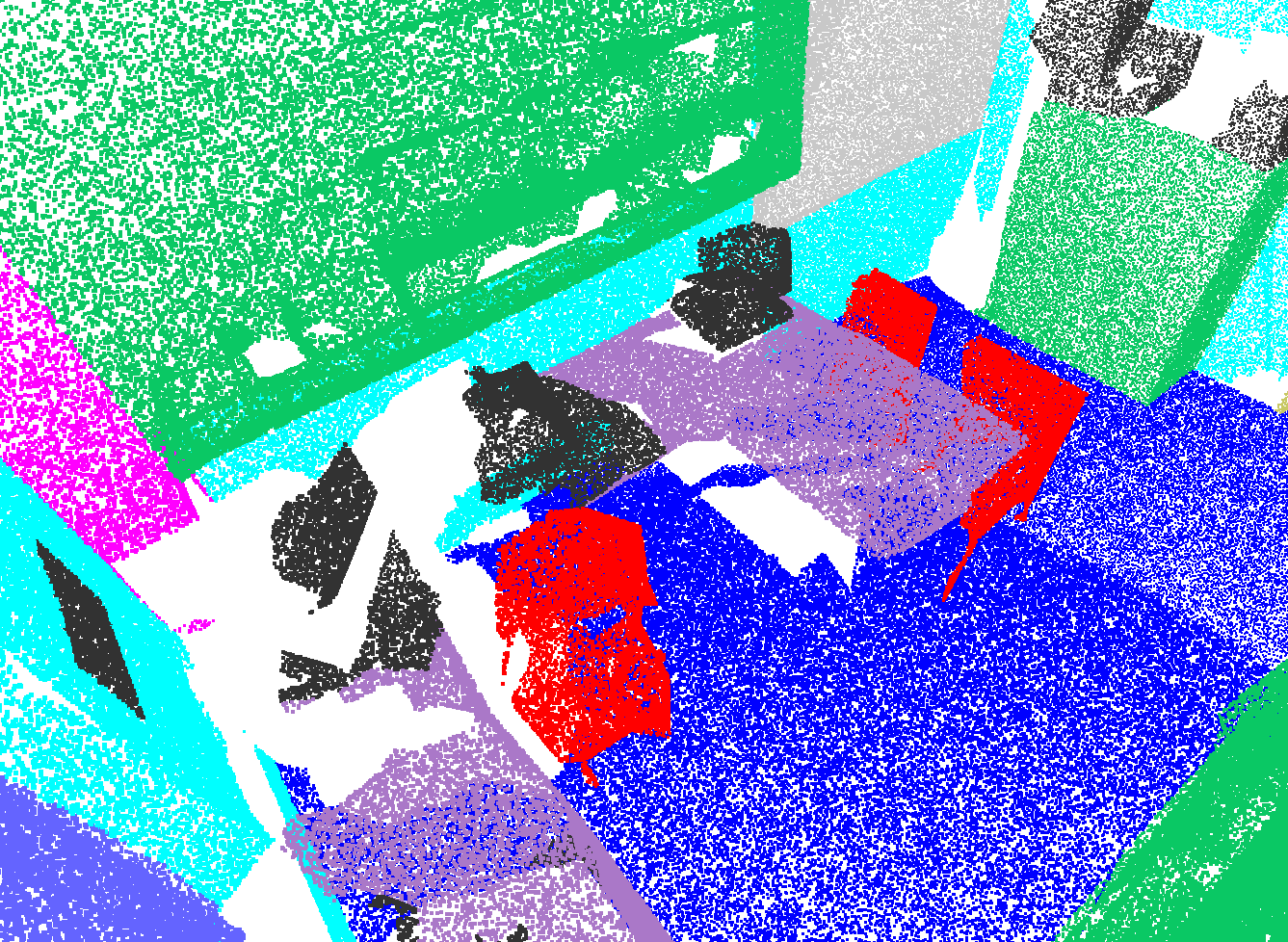}&
		\includegraphics[width=0.19\linewidth, height=2.3cm]{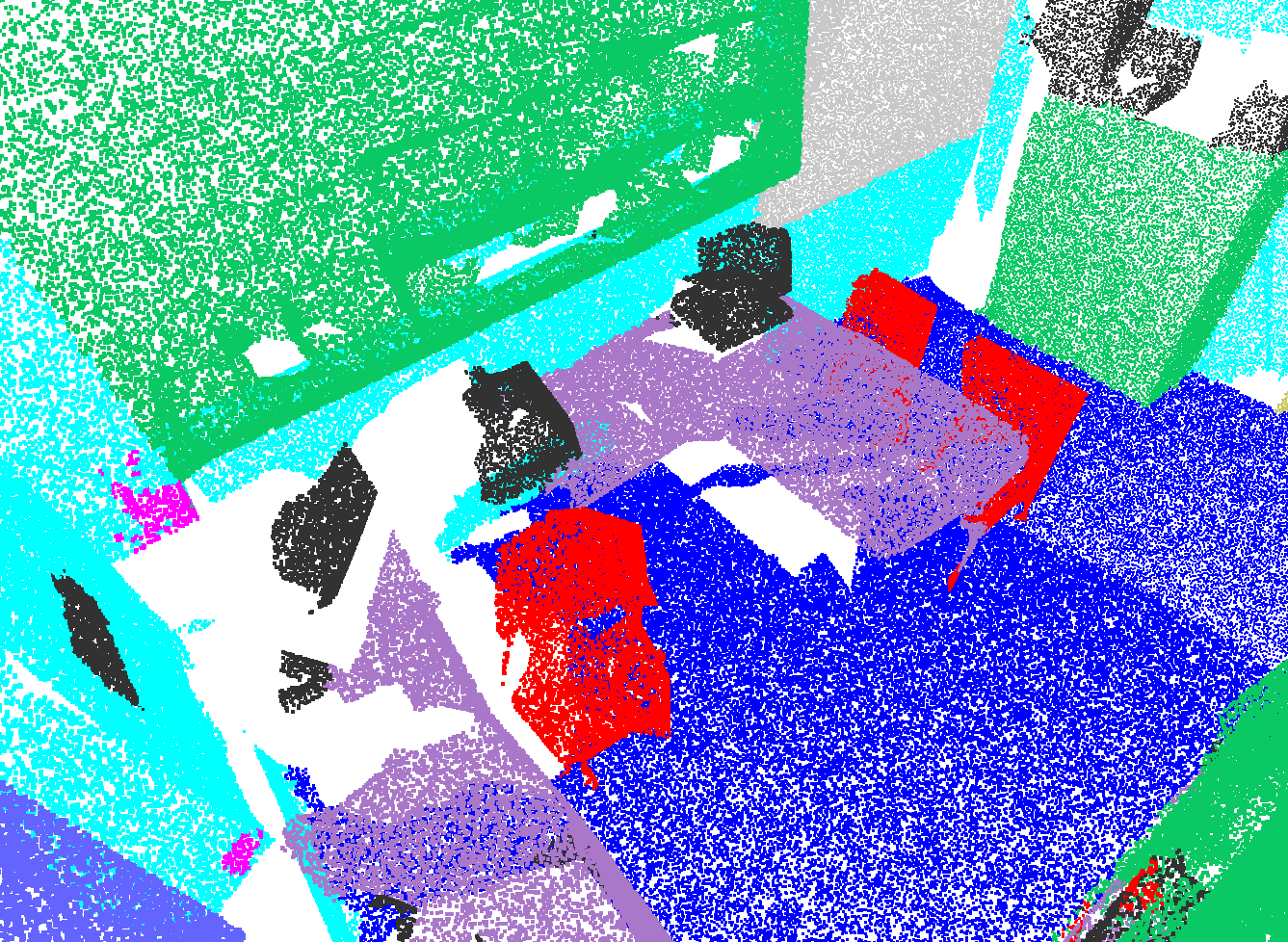}&
		\includegraphics[width=0.19\linewidth, height=2.3cm]{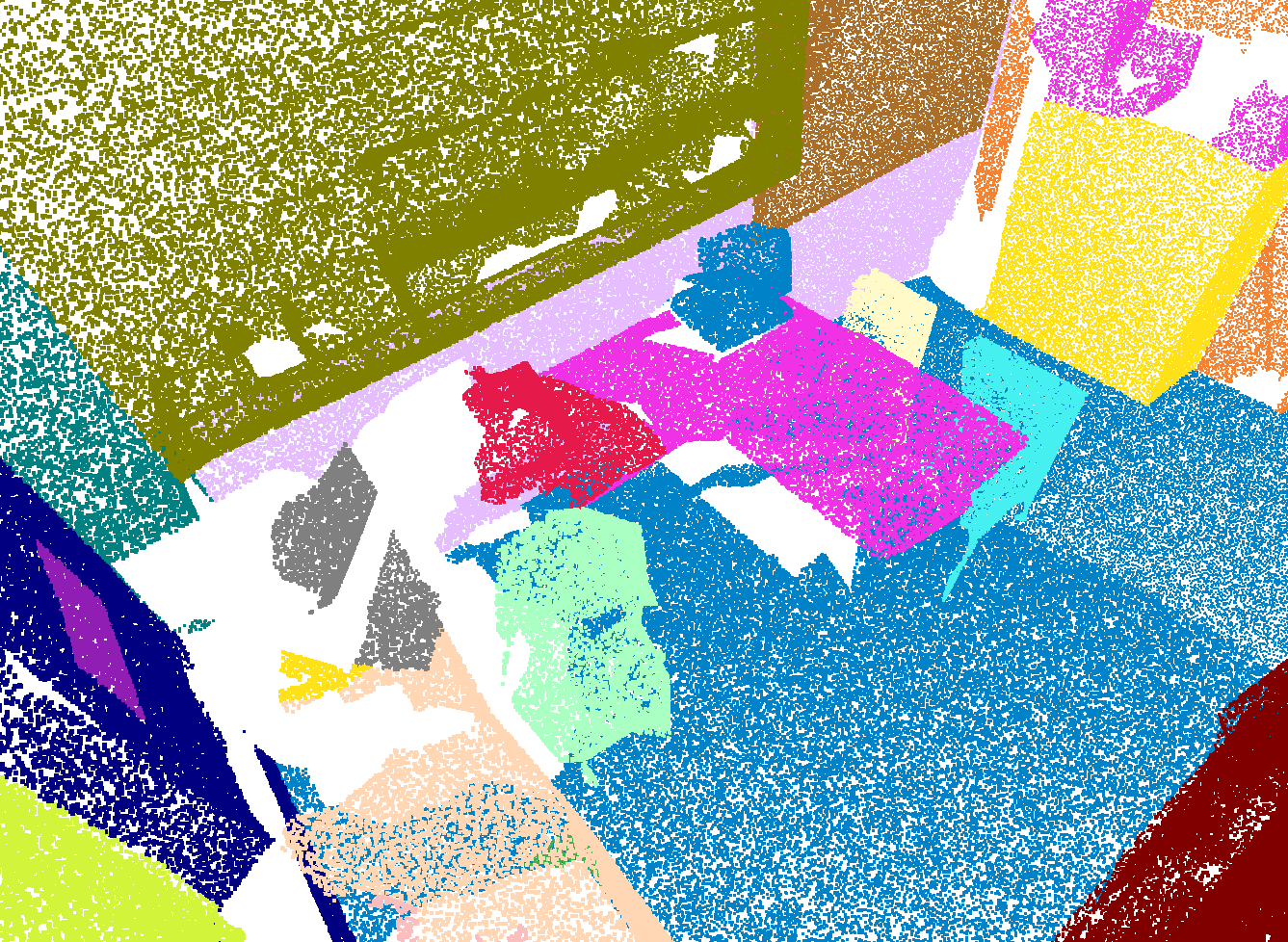}&
		\includegraphics[width=0.19\linewidth, height=2.3cm]{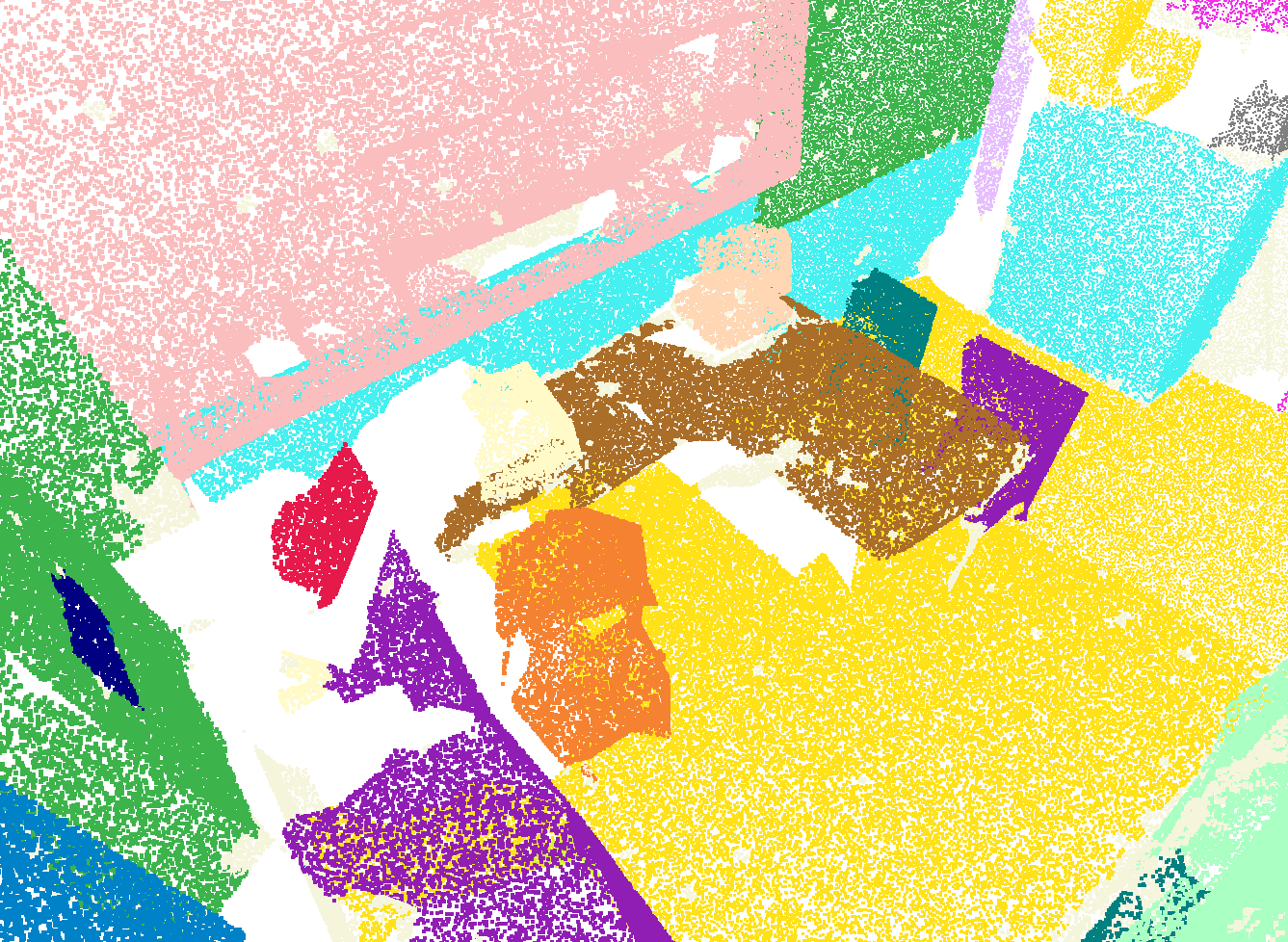}\\
		\includegraphics[width=0.19\linewidth, height=2.3cm]{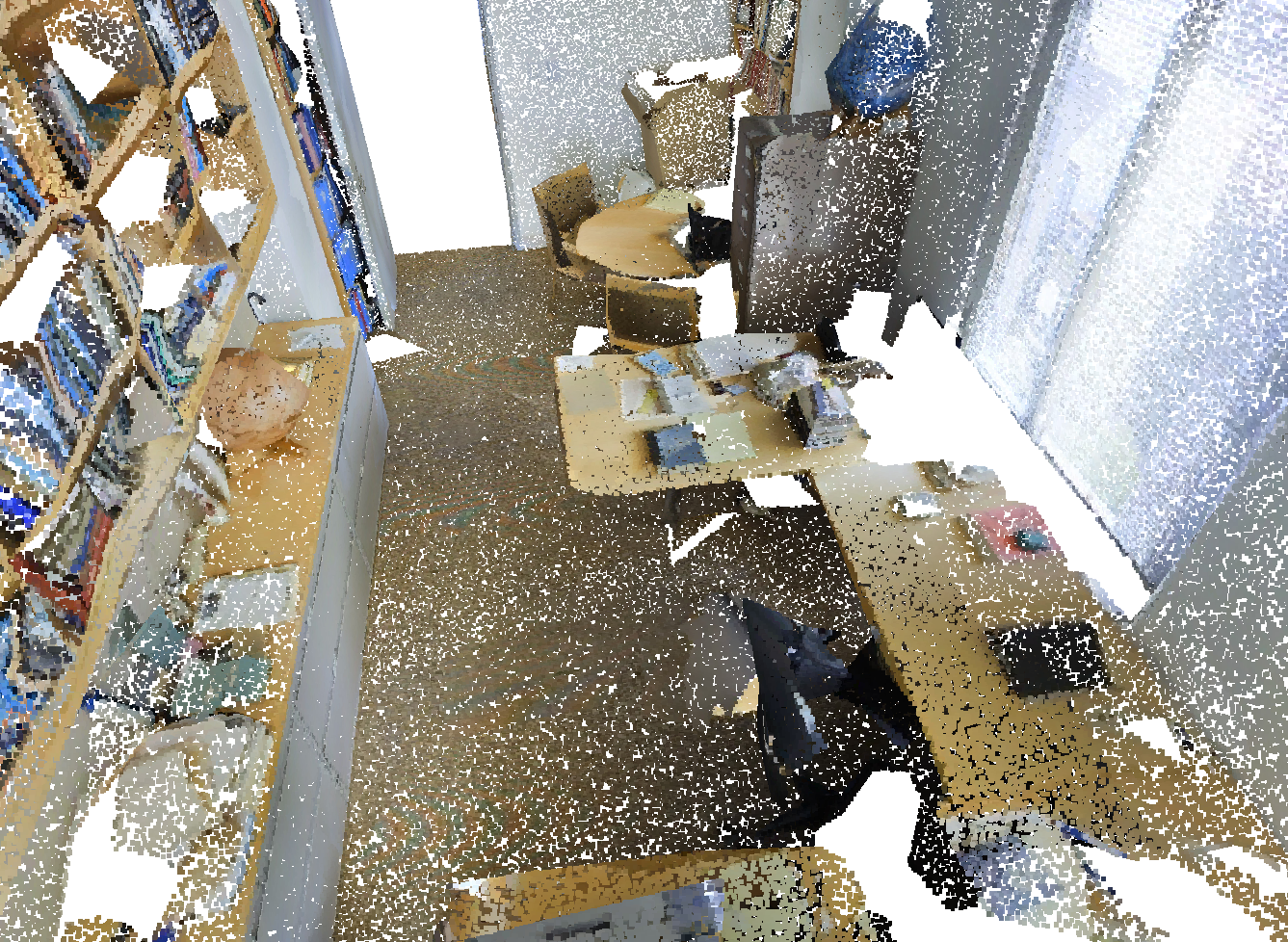}&
		\includegraphics[width=0.19\linewidth, height=2.3cm]{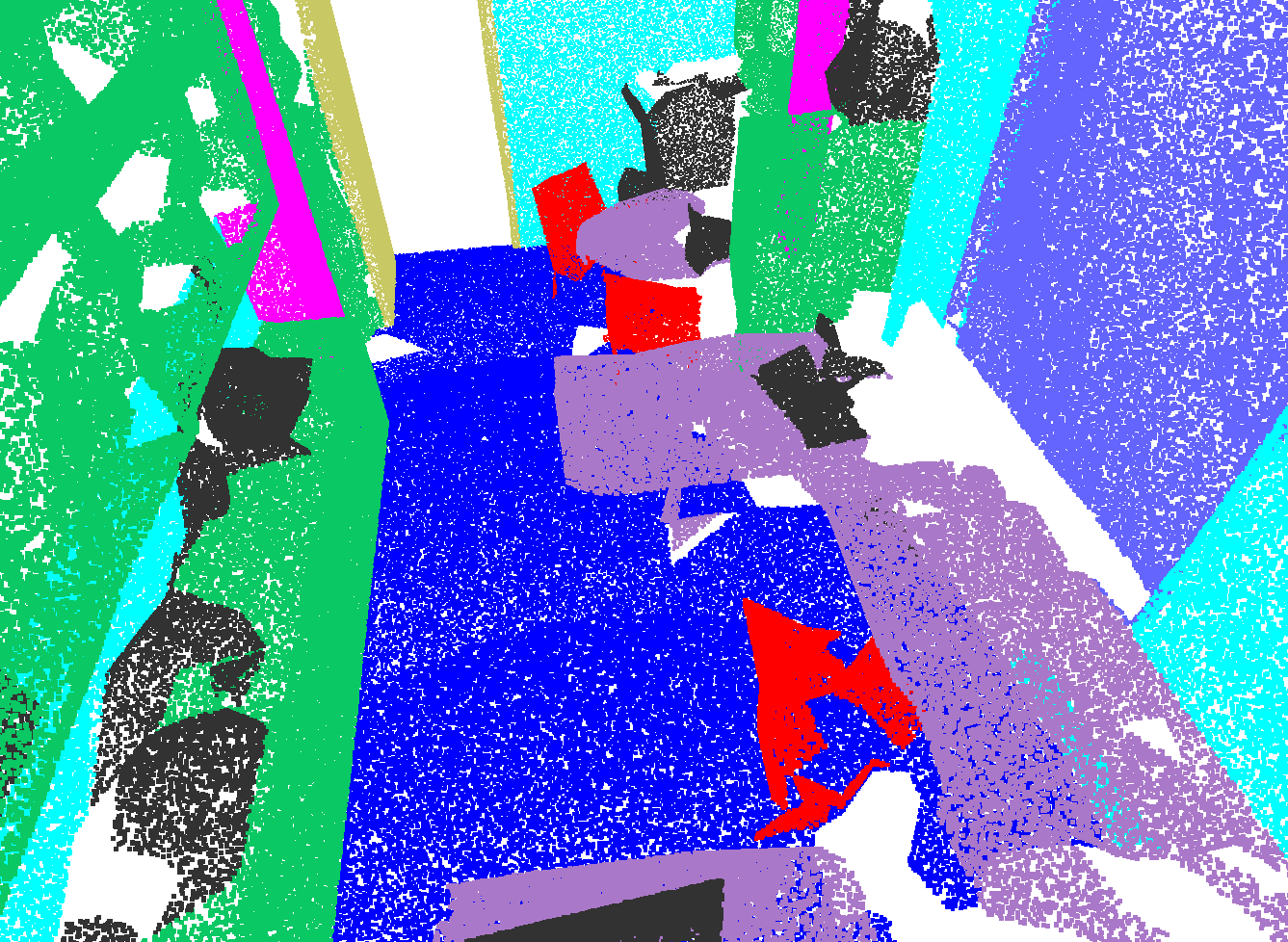}&
		\includegraphics[width=0.19\linewidth, height=2.3cm]{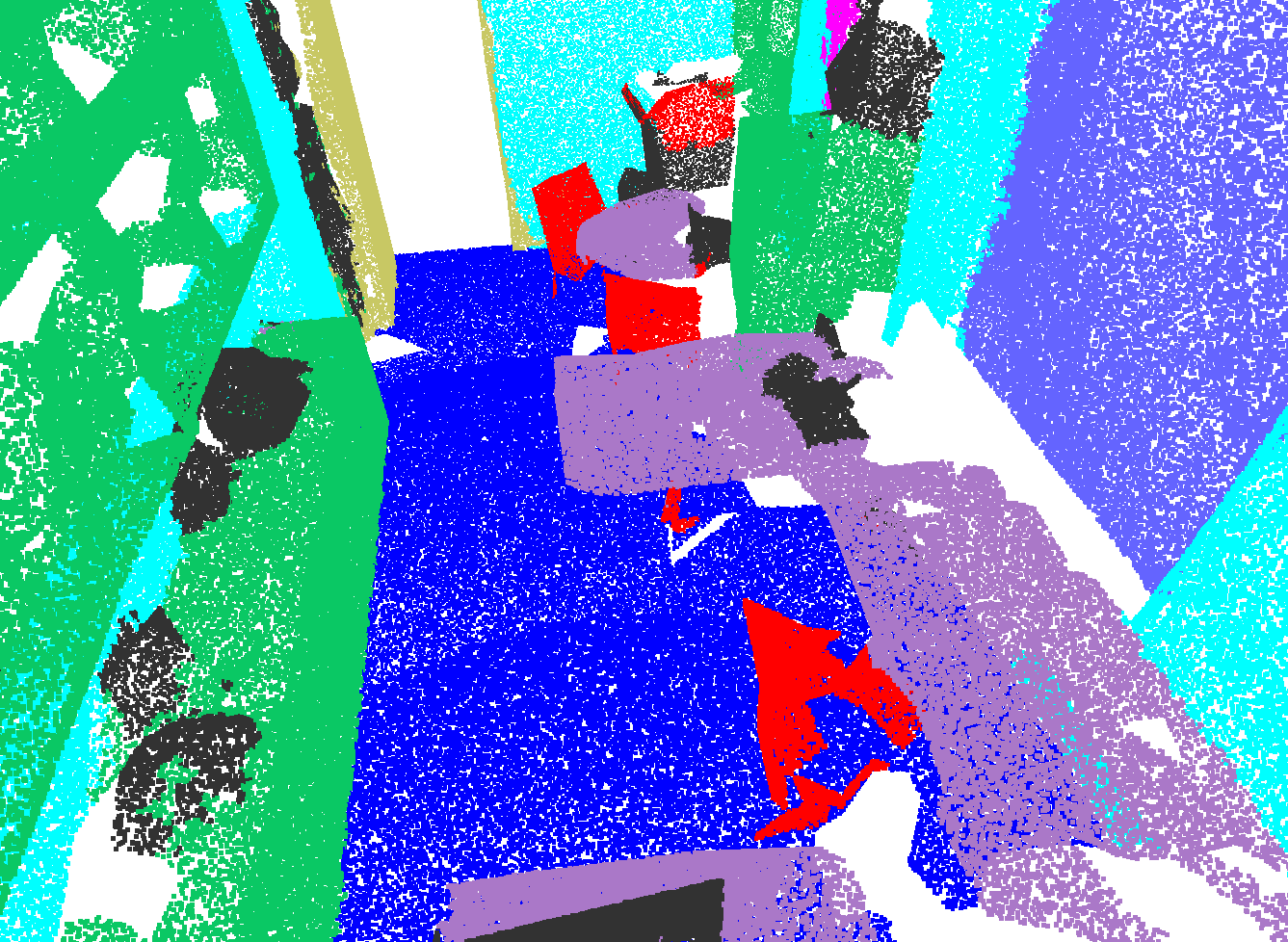}&
		\includegraphics[width=0.19\linewidth, height=2.3cm]{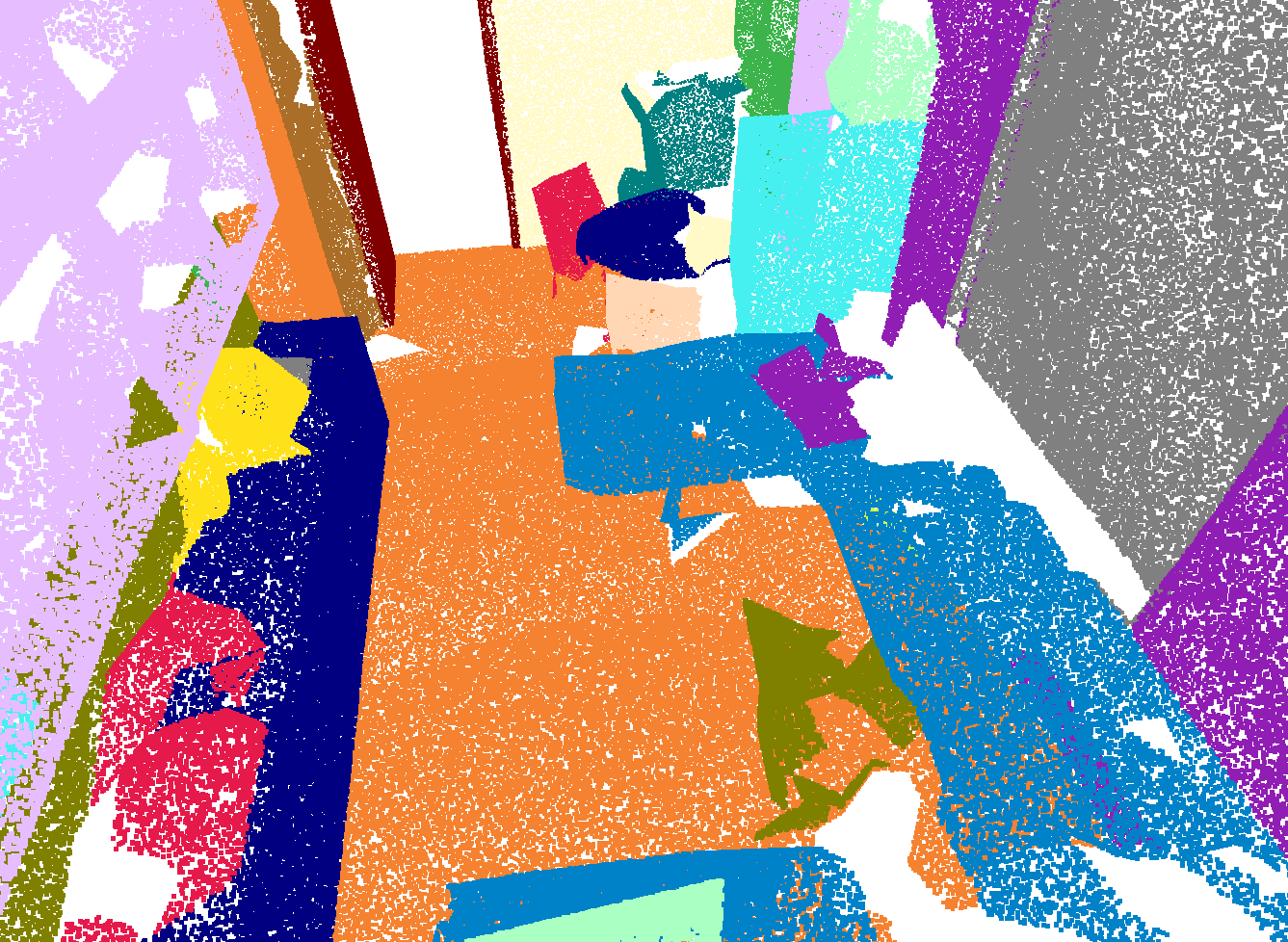}&
		\includegraphics[width=0.19\linewidth, height=2.3cm]{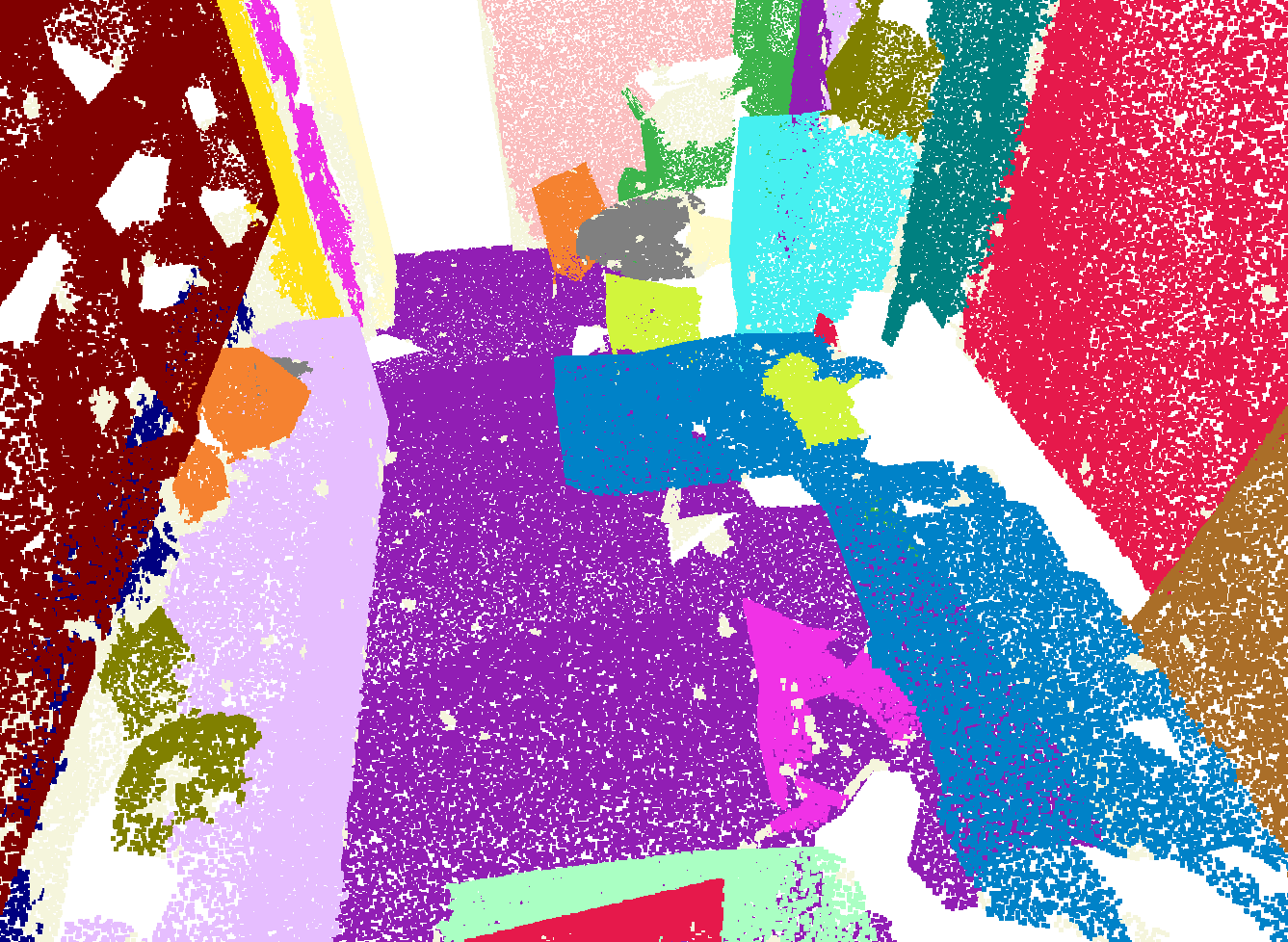}\\
		Input & Semantic GT & Semantic Pred. & Instance GT & Instance Pred.\\
	\end{tabular}
	\caption{Visualization of the semantic and instance segmentation results on ScanNet v2 (top) and S3DIS (bottom). For instance predictions, different colors represent separate instances, and the semantic results indicate the categories of instances.}
	\label{fig:visualization}
\end{figure*}

\subsection{Evaluation on S3DIS}
We also evaluate our proposed PointGroup model on the S3DIS dataset. 
Apart from adopting AP$_{50}$ as an evaluation metric, we also include the mPrec$_{50}$ and mRec$_{50}$ results in Table~\ref{tab:s3dis-compare}, where we use a score threshold of 0.2 to remove some low-confidence clusters. 

PointGroup reaches the highest performance in terms of all three evaluation metrics. For results on Area 5, PointGroup gets 57.8\% on AP$_{50}$, 61.9\% on mPrec$_{50}$ and 62.1\% on mRec$_{50}$. The mPrec$_{50}$ and mRec$_{50}$ are 6.6 and 19.7 points higher than ASIS~\cite{wang2019associatively}, respectively. For the results on 6-fold cross validation, PointGroup is 9.6 points higher than 
SGPN~\cite{wang2018sgpn} regarding AP$_{50}$, which is a big margin. 
The mPrec$_{50}$ and mRec$_{50}$ scores are 4 and 21.6 points higher than the second-best solution~\cite{yang2019learning}. 

The large improvement of PointGroup over the former best approaches across different challenging datasets demonstrate its effectiveness and generality. Several visual illustrations of PointGroup over these two datasets are included in Fig.~\ref{fig:visualization}. We observe that the proposed approach well captures the 3D geometry information and obtains precise instance segmentation masks.

\section{Conclusion}
We have proposed PointGroup for 3D instance segmentation, with a specific focus of better grouping points by exploring the in-between space and point semantic labels among the object instances.
Considering the situation that two intra-category objects may be very close to each other, we design a two-branch network to respectively learn a per-point semantic label and a per-point offset vector for moving each point towards its respective instance centroid. 
We then cluster points based on both the original point coordinates and the offset-shifted point coordinates. It combines the complementary strength of the two coordinate sets to optimize point grouping precision. Further, we introduced the ScoreNet to learn to evaluate the generated candidate clusters, followed by the NMS to avoid duplicates before we output the final predicted instances. 
PointGroup accomplished the best ever results. 

In our future work, we plan to further introduce a progressive refinement module to relieve the semantic inaccuracy problem that affects the instance grouping and explore the possibility of incorporating weakly- or self-supervision techniques to further boost the performance.

\vspace{-3.5mm}
\paragraph{Acknowledgments}
This project is supported in part by the Research Grants Council of the Hong Kong Special Administrative Region (Project no. CUHK 14201717).

{\small
\bibliographystyle{ieee_fullname}
\bibliography{egbib}
}

\end{document}